\documentclass[journal]{IEEEtran}
\usepackage{amsmath,amsfonts}
\usepackage{array}
\usepackage[caption=false,font=normalsize,labelfont=sf,textfont=sf]{subfig}
\usepackage{textcomp}
\usepackage{stfloats}
\usepackage{url}
\usepackage{verbatim}
\usepackage{graphicx}
\usepackage{cite}

\usepackage{booktabs}
\usepackage{bigstrut}
\usepackage{multirow}
\usepackage{makecell}
\usepackage[ruled]{algorithm2e}
\usepackage{verbatim}
\usepackage{url}
\usepackage{float}
\usepackage{color}
\usepackage{xcolor}
\usepackage[switch]{lineno}

\newcommand{\tabincell}[2]{\begin{tabular}{@{}#1@{}}#2\end{tabular}}

\begin{document}

\title{Self-Learning Symmetric Multi-view Probabilistic Clustering}

\author{Junjie Liu, Junlong Liu, Rongxin Jiang, Yaowu Chen, Chen Shen, Jieping Ye, ~\IEEEmembership{Fellow,~IEEE}
\thanks{Junjie Liu is with the College of Biomedical Engineering and Instrument Science, Zhejiang University and Alibaba Cloud, Hangzhou, China. (e-mail: jumptoliujj@gmail.com).}
\thanks{Junlong Liu, Chen Shen and Jieping Ye are with the Alibaba Cloud, Hangzhou, China. (e-mail: pingwu.ljl@alibaba-inc.com, jason.sc@alibaba-inc.com and yejieping.ye@alibaba-inc.com).}
\thanks{Rongxin Jiang is with the Zhejiang University and the Zhejiang Provincial Key Laboratory for Network Multimedia Technologies (e-mail: rongxinj@zju.edu.cn).}
\thanks{Yaowu Chen is with the Zhejiang University and the Zhejiang University Embedded System Engineering Research Center, Ministry of Education of China (e-mail: cyw@mail.bme.zju.edu.cn).}

\thanks{Corresponding author: Chen Shen (e-mail: jason.sc@alibaba-inc.com).}
\thanks{This work was done when Junjie Liu was a research intern at Alibaba.}
}

\markboth{IEEE TRANSACTIONS ON KNOWLEDGE AND DATA ENGINEERING}
{Shell \MakeLowercase{\textit{et al.}}: A Sample Article Using IEEEtran.cls for IEEE Journals}


\maketitle

\begin{abstract}
Multi-view Clustering (MVC) has achieved significant progress, with many efforts dedicated to learn knowledge from multiple views. However, most existing methods are either not applicable or require additional steps for incomplete MVC. Such a limitation results in poor-quality clustering performance and poor missing view adaptation. Besides, noise or outliers might significantly degrade the overall clustering performance, which are not handled well by most existing methods. In this paper, we propose a novel unified framework for incomplete and complete MVC named self-learning symmetric multi-view probabilistic clustering (SLS-MPC). SLS-MPC proposes a novel symmetric multi-view probability estimation and equivalently transforms multi-view pairwise posterior matching probability into composition of each view's individual distribution, which tolerates data missing and might extend to any number of views. Then, SLS-MPC proposes a novel self-learning probability function without any prior knowledge and hyper-parameters to learn each view's individual distribution. Next, graph-context-aware refinement with path propagation and co-neighbor propagation is used to refine pairwise probability, which alleviates the impact of noise and outliers. Finally, SLS-MPC proposes a probabilistic clustering algorithm to adjust clustering assignments by maximizing the joint probability iteratively without category information. Extensive experiments on multiple benchmarks show that SLS-MPC outperforms previous state-of-the-art methods.
\end{abstract}

\begin{IEEEkeywords}
Complete and Incomplete Multi-view Clustering, Multi-view Pairwise Posterior Matching Probability, Probabilistic Clustering, Probability Estimation and Refinement.
\end{IEEEkeywords}

\section{Introduction}
\IEEEPARstart{M}{ulti-view} clustering (MVC) \cite{MVC} aims at exploiting both correlated and complementary information from multi-view data and improving clustering performance beyond single-view clustering. With the explosion of multi-source and multi-modal data, a great deal of effort has been put into MVC. Different methods have been proposed to handle multi-view data, trying to classify samples into various clusters. Co-Regularization\cite{Co-Regularization}, based on co-training, intends to learn classifiers in each view through forms of multi-view regularization. Large-Scale Bipartite Graph\cite{Large-Scale-BG} fuses local manifold to integrate heterogeneous features and uses bipartite graphs to improve efficiency for large-scale MVC tasks. MKKM\cite{MKKM} proposes an effective matrix-induced regularization to enhance the diversity of the selected kernels, trying to maximize the kernel alignment. BMVC\cite{BMVC} first introduces a compact common binary code space for MVC task to optimize clusters in the hamming space with bit-operations. SMSC\cite{SMSC} seeks to learn the importance of different views and integrates anchor learning and graph construction into a unified framework to capture the complementary information from multiple views.

Despite previous progresses, MVC methods still face various challenges. Absence of partial views among data points \cite{EEIMC,Multi-SourceLearning} might frequently take place in practice, while existing methods are either not applicable \cite{GMC,SMSC} or require specific additional steps \cite{UEAF,PIC} for these cases. Such a limitation results in poor-quality clustering performance and poor missing view adaptation. Besides, noise or outliers might significantly degrade the overall clustering performance, which are not handled well by most existing methods. Moreover, K-means\cite{KMeans} clustering and spectral\cite{SPC} clustering are usually used for MVC tasks at the last step. Most of existing methods are less practical in real world cases because they have complex hyper-parameters and use extra information, including but not limited to the number of categories. This information plays an important role in their methods, and the absence of this information either causes their methods to fail or may degrade their clustering performance.


To address these issues, we propose a novel unified framework for incomplete and complete MVC named self-learning symmetric multi-view probabilistic clustering (SLS-MPC). It is difficult and complicated to learn a fusion similarity matrix in a linear or nonlinear manner based on the original similarity matrix. Thus, from a new perspective of probability, we utilize posterior probability to directly measure the probability that two samples belong to the same class. To obtain the posterior probability matrix, SLS-MPC mathematically decomposes it into the formulas of each views' distribution, which can extend to any number of views in an easy way. The proposed multi-view pairwise posterior matching probability is symmetric for each view and tolerates view missing in an intuitive way. Then, equipped with the consistency information excavation in single-view, cross-view and multi-view, a novel self-learning probability function is proposed to effectively learn each view's individual distribution without any prior knowledge and hyper-parameters. Next, SLS-MPC performs graph-context-aware probability refinement with path propagation and co-neighbor propagation, which can effectively alleviate the impact of noise and outliers. Finally, clusters are generated using the proposed probabilistic clustering algorithm, which is more robustness to noise and does not require the prior knowledge of cluster numbers. Extensive experiments demonstrate that SLS-MPC significantly outperforms state-of-the-art methods.

In summary, the main novelties of this paper are as follows:
\begin{itemize}
  \item A novel symmetric pairwise posterior matching probability is proposed and SLS-MPC equivalently transforms multi-view pairwise posterior matching probability into compositions of each view's individual distribution, which tolerates data missing and might extend to any number of views.
  \item To fully dig out the consistency information from multiple views in an unsupervised manner, a novel self-learning probability function is proposed to effectively learn each view's individual distribution without any prior knowledge and hyper-parameters.
  \item To further alleviate the impact of noise and outliers, a novel graph-context-aware refinement is proposed based on the aspect of graph context.
  \item Besides, a novel probabilistic clustering algorithm is proposed to generate clustering results in an unsupervised manner without any prior knowledge.
  \item Extensive experiments on multiple benchmarks for incomplete and complete MVC show that SLS-MPC significantly outperforms previous state-of-the-art methods.
\end{itemize}

\section{Related Work}
\label{sec:relatedwork}
A modern MVC method is usually composed of two parts, a consistent representation constructed from all views which is used to learn consensus from multi-view data and a clustering algorithm based on the consistent representation which is used to generate clustering result. Based on the mechanisms and principles used in learning consensus from multiple views, existing MVC algorithms can be grouped into several categories. The first category is based on graph clustering\cite{GH-W,GH-OS,GMC,PIC}. As a typical graph clustering method, PIC\cite{PIC} seeks to complete the similarity matrix and learn a consensus matrix and finally performs spectral clustering on the consensus laplacian matrix. GMC\cite{GMC} weights each view's graph matrix to learn a unified graph matrix. The second one is based on matrix factorization\cite{MF,MF-J,MF-L2,MF-R,SC_low_rank_mat,OC_beyond_low_rank}. This category seeks to learn a consensus representation by performing low-rankness to achieve clustering. For example, MIC\cite{MF-L2}, based on weighted non-negative matrix factorization and $L_{2,1}$-norm regularization, minimizes the consensus by learning the latent feature matrices for each view. The third one is multiple kernel learning\cite{MKC,MKC-LF,MKC-NK,OSLF}. In brief, this category seeks to combine different predefined kernels either linearly or non-linearly in order to arrive at a unified kernel. For example, OSLF\cite{OSLF} proposes to learn consensus cluster partition matrix by combing linearly-transformed base partitions obtained from single views. Besides, the methods like \cite{DMVC-AE2,MCDCF,IMCCP} are based on deep multi-view clustering and MCDCF\cite{MCDCF} performs multi-layer concept factorization and derives a common consensus representation matrix from the hierarchical information. Moreover, some ensemble-based\cite{EnsembleC} MVC methods and scalable\cite{SMSC,SFMC} MVC methods are proposed to advance MVC understanding in new ways. Different from the aforementioned methods, we propose a novel self-learning probability function to effectively learn each view's individual distribution without any prior knowledge and hyper-parameters from the aspect of consistency in single-view, cross-view and multi-view and a novel method to adaptively estimate the posterior matching probability from multiple views without complicated hyper-parameters fine-tuning.

K-means clustering\cite{KMeans}, spectral clustering\cite{SPC}, hierarchical clustering\cite{HAC} and some other traditional clustering algorithms\cite{AP,DBSCAN} are usually used for clustering tasks. With a given number of clusters $K$, K-means clustering\cite{KMeans} is an iterative algorithm that tries to partition samples into $K$ clusters and makes the intra-cluster data points as similar as possible while also keeping the clusters as far as possible by minimizing the total intra-cluster variance. Spectral clustering\cite{SPC} uses information from the eigenvalues of similarity matrix derived from the graph and seeks to choose appropriate eigenvectors to cluster different data points. Hierarchical clustering\cite{HAC} seeks to create a hierarchical clustering tree in which the original data is at the bottom and the root node is at the top. The clustering performance of these algorithms is affected by the optimization parameters and the number of clusters. As one of effective clustering algorithms, probabilistic clustering algorithms\cite{SLPPC, PPC} are pioneered to incorporate pairwise relations and have achieved state-of-the-art performance in clustering tasks. The basic idea of probabilistic clustering is to maximize the intra-cluster similarities and minimize the inter-cluster similarities among the objects. Empirical functions and weighted confidence or preference are usually used to separate samples, which limits the final clustering performance. Moreover, the matching probability of all pairwise relations are taken into consideration in \cite{SLPPC, PPC} resulting in high computational complexity. Besides, the number of categories is used in optimization process in some methods and these information plays an important role in their methods\cite{PIC,OSLF}, without which either causing the failure of their methods or might degrade the performance. Thus, we propose a novel probabilistic clustering algorithm, which has no optimization parameters and generates clustering results in an unsupervised manner and an efficient way without category information.

This work is different from existing methodologies in several key aspects. First, almost all these methods\cite{SC_low_rank_mat,OC_beyond_low_rank,UEAF,GMC,PIC,OSLF,MCDCF,IMCCP,SMSC,SFMC} contain complicated model design, which make them infeasible in real-world applications. In contrast, our SLS-MPC contains an intuitive and efficient clustering framework with multiple clear steps, including symmetric multi-view probability estimation, probability function self-learning, graph-context-aware refinement and probabilistic clustering. Second, different from the works like \cite{GMC,PIC,UEAF}, our SLS-MPC seeks to adaptively handle multi-view data and missing data from a probabilistic perspective rather than fusing multi-view data using a set of weights, thus embracing higher explainability. In addition, this paper is extended from MPC\cite{MPC} but differs in the following two aspects. First, multi-view probability estimation has been optimized from an asymmetric form to a symmetric form (Section \ref{sec:sub_mpe}). This advancement eliminates the inherent issue of view order selection in the asymmetric form in MPC and ensures consistency in the probability form across all views. Second, MPC utilizes pseudo-labels to independently estimate each view's probability function. However, pseudo-labels may conflict across different views, making it difficult to ensure the consistency between the estimated probability functions. In contrast, our method proposes a novel self-learning probability function (Section \ref{sec:sub_selflearning}) to effectively learn each view's individual distribution from the perspective of consistency of probability function. The proposed self-learning probability function, in conjunction with the other components of our method, constitutes a more robust theoretical framework.

\section{Methodology}
\label{sec:method}
\subsection{Symmetric Multi-view Probability Estimation}
\label{sec:sub_mpe}
Given a multi-view dataset of $N$ samples with $M$ views $S=\{V^{(1)},V^{(2)},...,V^{(M)}\}$. $V^{(m)} \in R^{d^{(m)}*N}$ denotes the feature matrix in $m$-th view, where $d^{(m)}$ is the feature dimension of the $m$-th view. Let $W^{(m)} \in R^{N*N}$ calculated by $V^{(m)}$ using cosine similarity denotes the similarity matrix of the $m$-th view. Assuming that all views are conditionally independent similar to previous works\cite{Indep-Bootstrapping,Indep-Subspace,Indep-PSubspace,Indep-MVC,Indep-Correlation}, the pairwise posterior probability of sample $i$ and $j$ proposed in MPC\cite{MPC} is: 
\begin{equation}\label{equ:p_bayesian}
\begin{split}
    P(i,j) &= P(e_{ij}=1|w^{(1)}_{ij},w^{(2)}_{ij},...,w^{(M)}_{ij}) \\
    & = \frac {(\prod \limits_{m=2}^M P(w^{(m)}_{ij}|e_{ij}=1))P(e_{ij}=1|w^{(1)}_{ij})} {\sum \limits_{l \in\{0,1\}} (\prod \limits_{m=2}^M P(w^{(m)}_{ij}|e_{ij}=l))P(e_{ij}=l|w^{(1)}_{ij})} \\
\end{split}
\end{equation}
where $e_{ij}$ indicates that the two samples belong to the same class and $w^{(m)}_{ij}$ denotes the similarity of the two samples in $m$-th view. Eq. (\ref{equ:p_bayesian}) is asymmetric for each view and has two types of probability function. Considering the consistent representation across multiple views, we further derive the Eq. (\ref{equ:p_bayesian}). Let $d_m=w^{(m)}_{ij}, e_1=(e_{ij}=1), e_0=(e_{ij}=0)$ for short and Eq. (\ref{equ:p_bayesian}) can be expressed as:
\begin{equation}\label{equ:p_vv}
\begin{split}
    P(i,j) &= P(e_1|d_1,d_2,...,d_M) \\
    &= \frac {(\prod \limits_{m=2}^M P(d_m|e_1))P(e_1|d_1)} {\sum \limits_{e \in \{e_0,e_1\}} (\prod \limits_{m=2}^M P(d_m|e))P(e|d_1)} \\
\end{split}
\end{equation}
Based on Bayesian formula, $P(d_m|e_1)$ and $P(d_m|e_0)$ can be expressed as:
\begin{equation}\label{equ:p_ve}
\begin{split}
    P(d_m|e_1) = \frac {P(e_1|d_m)P(d_m)} {P(e_1)} \\
    P(d_m|e_0) = \frac {P(e_0|d_m)P(d_m)} {P(e_0)} \\
\end{split}
\end{equation}
Naturally, Eq. (\ref{equ:p_vv}) can be expressed as:
\begin{equation}\label{equ:p_vf}
\begin{split}
    P(i,j) &= P(e_1|d_1,d_2,...,d_M) \\
    &= \frac {(\prod \limits_{m=2}^M \frac {P(e_1|d_m)P(d_m)} {P(e_1)})P(e_1|d_1)} {\sum \limits_{l \in \{0,1\}} (\prod \limits_{m=2}^M \frac {P(e_l|d_m)P(d_m)} {P(e_l)})P(e_l|d_1)} \\
    &= \frac {(\prod \limits_{m=1}^M P(e_1|d_m)) P(e_0)^{M-1}} {\sum \limits_{l \in \{0,1\}} (\prod \limits_{m=1}^M P(e_l|d_m)) P(e_{1-l})^{M-1}}
\end{split}
\end{equation}
Thus, the pairwise probability of sample $i$ and $j$ can be expressed as:
\begin{small}
\begin{equation}\label{equ:p_bayesian_tmp}
    P(i,j) = \frac {(\prod \limits_{m=1}^M P(e_{ij}=1|w^{(m)}_{ij}))P_0} {(\prod \limits_{m=1}^M P(e_{ij}=1|w^{(m)}_{ij}))P_0+(\prod \limits_{m=1}^M P(e_{ij}=0|w^{(m)}_{ij}))P_1}
\end{equation}
\end{small}
where $P_0=P(e_{ij}=0)^{M-1}$ and $P_1=P(e_{ij}=1)^{M-1}$. Given sample $i$ and sample $j$ without any prior information, the two samples either belong to the same class or do not belong to the same class, which indicates $P(e_{ij}=0)=P(e_{ij}=1)=0.5$.
Finally, the pairwise probability of sample $i$ and $j$ can be expressed as:
\begin{small}
\begin{equation}\label{equ:p_bayesian_new}
    P(i,j) = \frac {\prod \limits_{m=1}^M P(e_{ij}=1|w^{(m)}_{ij})} {\prod \limits_{m=1}^M P(e_{ij}=1|w^{(m)}_{ij})+\prod \limits_{m=1}^M P(e_{ij}=0|w^{(m)}_{ij})}
\end{equation}
\end{small}
which is symmetric for each view.

\subsection{Self-Learning Probability Function}
\label{sec:sub_selflearning}
\begin{figure*}
    \centering
    \includegraphics[width=0.98\textwidth]{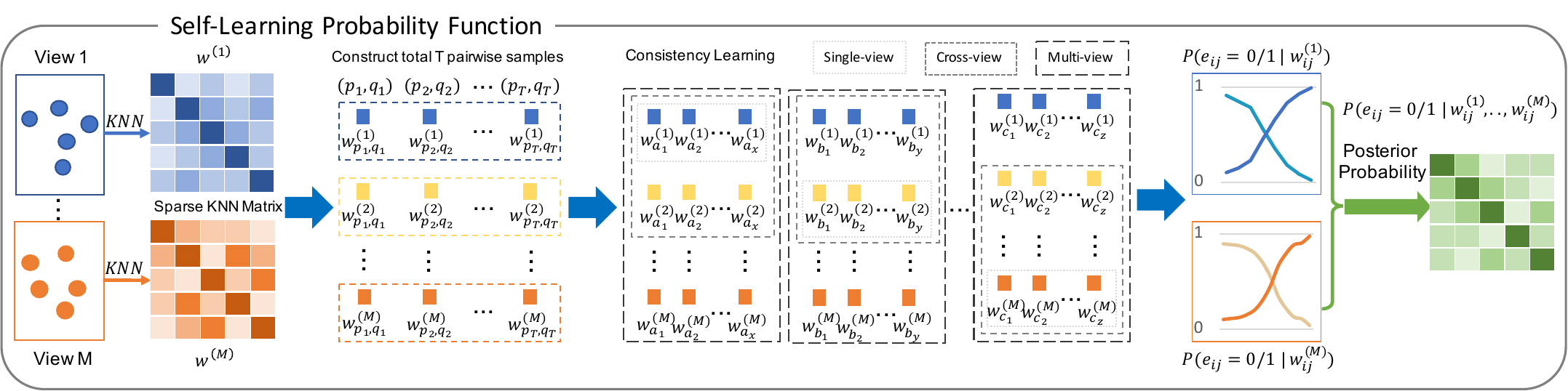}
    \caption{Illustration of the self-learning probability function. Given a multi-view dataset of $N$ samples with $M$ views $S=\{V^{(1)},V^{(2)},...,V^{(M)}\}$, $KNN^{(m)} \in R^{N*K}$ can be generated on the similarity matrix $W^{(m)} \in R^{N*N}$ of the $m$-th view. $KNN^{(m)}$ construct the training data including total $T$ pairwise samples $(p_t,q_t)$ and the corresponding similarity values $(w^{(1)}_{p_t,q_t},w^{(2)}_{p_t,q_t},...,w^{(M)}_{p_t,q_t})$. We divide each view's data $\{w^{(m)}_{p_t,q_t}\}$ of total $T$ length into $I$ parts in the order of $\{w^{(m)}_{p_t,q_t}\}$ from small to large defined in Eq. (\ref{equ:f_function}). $a$, $b$ and $c$ are three specific parts in the total of $I$ parts from three specific views. The light gray dotted boxes represent the single-view forms from different views. The dark gray dotted box represent the cross-view forms from different views. And the black dotted box represent the multi-view forms from different views. The single-view, cross-view and multi-view probability functions are defined in Eq. (\ref{equ:f_single}), Eq. (\ref{equ:f_cross}) and Eq. (\ref{equ:f_multi}) and the consistency constraint is defined in Eq. (\ref{equ:loss_1}). Then a self-learning probability function is proposed to learn the $P(e_{ij}=1|w^{(m)}_{ij})$ from the aspect of consistency in single-view, cross-view and multi-view without any prior knowledge and hyper-parameters. Finally, a multi-view pairwise posterior matching probability matrix is generated from the composition of each view's individual distribution.}
    \label{fig:selflearning}
\end{figure*}


Eq. (\ref{equ:p_bayesian_new}) defines the decomposition form and the probability function $P(e_{ij}=1|w^{(m)}_{ij})$ for each view needs to be estimated. A simple way to estimate the probability function is using isotonic regression to fit the pairwise relationship between samples based on pseudo labels (pseudo labels can be generated on each view by a simple clustering algorithm, such as K-means). The performance of the MVC task depends on the quality of the generated pseudo labels. Besides, this simple approach estimates the probability function on each single view, overlooking the important consistent information across multiple views. Thus, to fully dig out the consistency information from multiple views in an unsupervised manner, we propose a self-learning probability function to learn the $P(e_{ij}=1|w^{(m)}_{ij})$ from the aspect of consistency in single-view, cross-view and multi-view without any prior knowledge and hyper-parameters. Fig. \ref{fig:selflearning} illustrates the detailed learning process of the proposed self-learning probability function. And, this section is structured as follows: (1) In Section \ref{sec:sub_function1}, we first introduce the motivation behind the self-learning probability function and provide the definition of consistency. (2) Section \ref{sec:sub_function2} presents the definitions of multiple probability functions that need to be used in consistency learning defined in the first step. (3) In Section \ref{sec:sub_function3}, we finally design the objective function to learn each view's individual distribution based on the definitions of consistency and multiple probability functions.

\begin{figure}[!ht]
    \centering
    \includegraphics[width=0.48\textwidth]{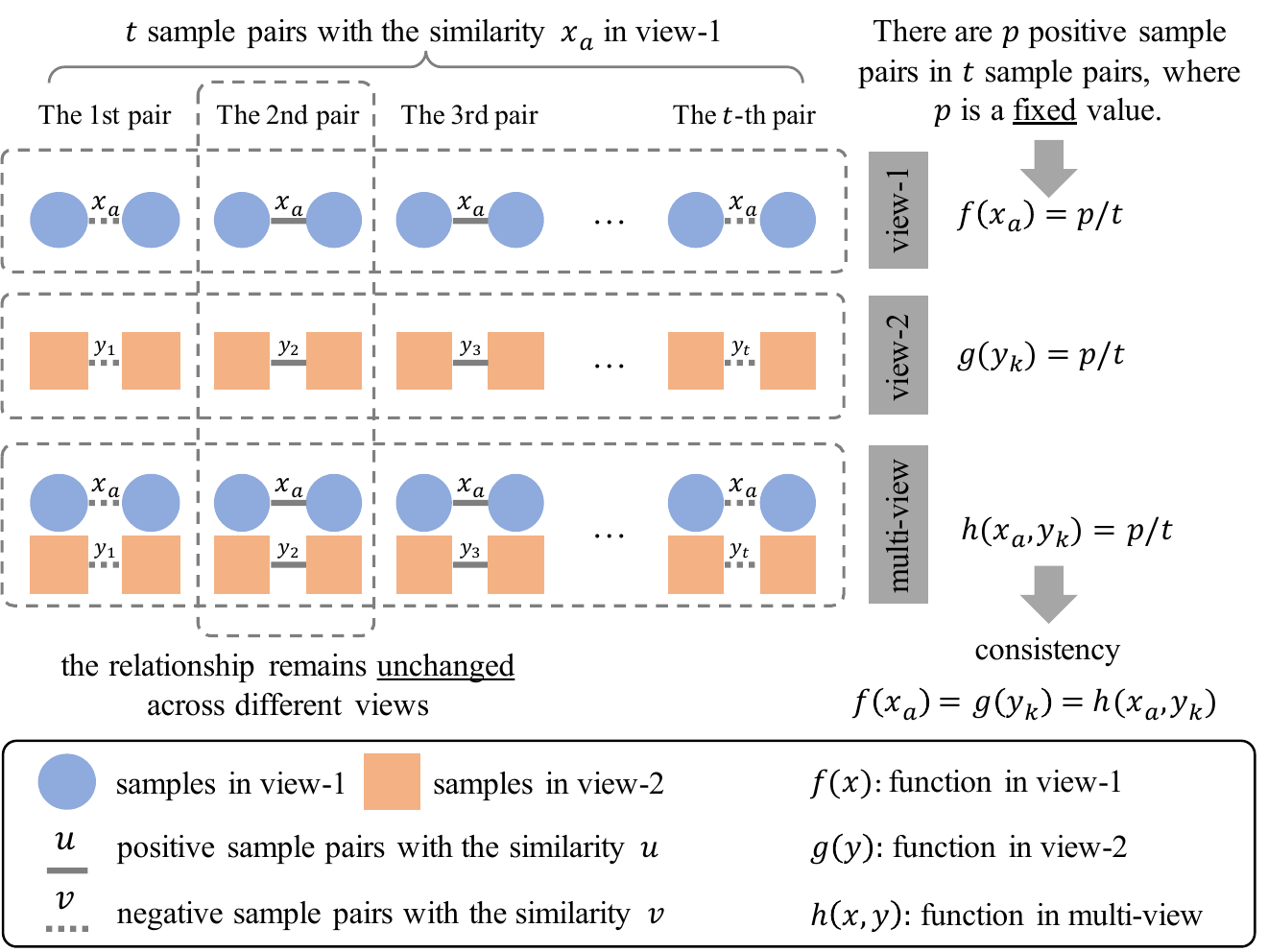}
    \caption{Our basic observation and motivation of self-learning probability function. Given the condition that the original similarity between the sample pairs in the first view is $x_a$, there are $t$ sample pairs including fixed $p$ positive sample pairs. Fix these $t$ sample pairs and find the original similarity between the sample pairs in the second view ($\{y_k|k \in \{1,...,t\}\}$). Due to the fixed sample pairs, the probability that the sample pairs belong to the same class in the first view ($f(x_a)$) and the second view ($g(y_k)$) should be consistent. In the same way, the probability that the sample pairs belong to the same class in the first view ($f(x_a)$) and multi-view ($h(x_a,y_k)$) should be also consistent.}
    \label{fig:xx}
\end{figure}

\subsubsection{{\textbf{Consistency Motivation and Definition}}}
\label{sec:sub_function1}
Firstly, we introduce the motivation behind the self-learning probability function. Taking two views as an example, we define the first view $P(e_{ij}=1|w^{(1)}_{ij})$ as a continuous monotonic function $f(x)$ as:
\begin{equation}\label{equ:functional_single}
\begin{split}
    f(x): P(e_{ij}=&1|w^{(1)}_{ij}=x) \\
    s.t. \ f(x_1) \leq f(&x_2), x_1 < x_2, \\
    f(x_{min}) = 0,\ &f(x_{max}) = 1 \\
\end{split}
\end{equation}
where $x \in \{w^{(1)}_{i,j}\}$, $x_{min} = min(w^{(1)}_{i,j})$ and $x_{max} = max(w^{(1)}_{i,j})$. In the same way, we define the second view $P(e_{ij}=1|w^{(2)}_{ij})$ as a continuous monotonic function $g(y)$ and $g(y)$ has the same constraints as $f(x)$ including range and monotonicity. And, the multi-view function $h(x,y)$ based on Eq. (\ref{equ:p_bayesian_new}) is defined as:
\begin{equation}\label{equ:functional_multi}
\begin{split}
    h(x,y) = \frac {f(x)g(y)} {f(x)g(y)+(1-f(x))(1-g(y))}
\end{split}
\end{equation}

As illustrated in Fig. \ref{fig:xx}, $f(x_{a})$ indicates the probability that the sample pairs belong to the same class given the similarity $x_a$ in the first view. A subset of pairwise samples $s=\{(i,j)|w^{(1)}_{ij}=x_a\}$ contains all pairs of samples with similarity $x_a$ in the first view and the proportion of pairwise samples of the same class in the subset $s$ is a fixed value. Then from the perspective of the second view, $\{(f(x_a),g(y_{k}))|y_{k}=w^{(2)}_{ij},(i,j)\in s\}$ contains the probability that the sample pairs belong to the same class in the subset $s$ from the first view and second view. Due to the fixed number of positive sample pairs in the subset $s$, the probability that the sample pairs belong to the same class in the first view and the second view should be consistent. Thus, we present the cross-view consistency as follows.

\textbf{{\textit{Definition 1:}}} The cross-view consistency from the first view ($f(x)$) to the second view ($g(y)$) can be mathematically expressed as:
\begin{equation}\label{equ:functional_con_fg}
\begin{split}
    L_{f-g} &= D(f(x), \frac {\int g(y)p(x,y)dy} {\int p(x,y)dy}) \\
\end{split}
\end{equation}
where $D$ is the distance function and $p(x,y)$ is the similarity distribution between the first view and second view. 

\textbf{{\textit{Definition 2:}}} The cross-view consistency from the second view ($g(y)$) to the first view ($f(x)$) can be expressed as:
\begin{equation}\label{equ:functional_con_gf}
\begin{split}
    L_{g-f} &= D(g(y), \frac {\int f(x)p(x,y)dx} {\int p(x,y)dx}) \\
\end{split}
\end{equation}

Furthermore, $\{(f(x_a),h(x_a,y_{k}))|y_{k}=w^{(2)}_{ij},(i,j)\in s\}$ contains the probability that the sample pairs belong to the same class in the subset $s$ from the perspective of multi-view. As illustrated in Fig. \ref{fig:xx}, the probability that the sample pairs belong to the same class in single-view and multi-view should be also consistent. Thus, we present the multi-view consistency as follows.

\textbf{{\textit{Definition 3:}}} The multi-view consistency from the single-view ($f(x)$ and $g(y)$) to multi-view ($h(x,y)$) can be mathematically expressed as:
\begin{equation}\label{equ:functional_fgh}
\begin{split}
    L_{f-h} &= D(f(x), \frac {\int h(x,y)p(x,y)dy} {\int p(x,y)dy}) \\
    L_{g-h} &= D(g(y), \frac {\int h(x,y)p(x,y)dx} {\int p(x,y)dx}) \\
\end{split}
\end{equation}

Finally, based on \textit{Definition 1-3}, we present the consistency constraint as follows.

\textbf{{\textit{Definition 4:}}} To constraint the function $f(x)$ and $g(y)$, consistency constraint can be mathematically expressed as:
\begin{equation}\label{equ:functional_con}
\begin{split}
    L_{consistency} &= L_{f-g} + L_{g-f} + L_{f-h} + L_{g-h} \\
    f,g &= \mathop{\arg\min}\limits_{f,g} L_{consistency}
\end{split}
\end{equation}
where $L_{f-g}$ and $L_{g-f}$ are the cross-view constraints, $L_{f-h}$ and $L_{g-h}$ are the multi-view constraints.

\subsubsection{{\textbf{Definitions of Probability Functions}}}
\label{sec:sub_function2}
Next, we present definitions of multiple probability functions that need to be used in the above consistency definition. As mentioned in Section \ref{sec:sub_mpe}, given a multi-view dataset of $N$ samples with $M$ views $S=\{V^{(1)},V^{(2)},...,V^{(M)}\}$, $KNN^{(m)} \in R^{N*K}$ can be generated on the similarity matrix $W^{(m)} \in R^{N*N}$ of the $m$-th view. Then $KNN^{(m)}$ construct the training data including total $T$ pairwise samples $(p_t,q_t)$ and the corresponding similarity values $(w^{(1)}_{p_t,q_t},w^{(2)}_{p_t,q_t},...,w^{(M)}_{p_t,q_t})$, $t=1,2,...,T$. Due to the complexity of solution in Eq. (\ref{equ:functional_con_fg}), Eq. (\ref{equ:functional_con_gf}) and Eq. (\ref{equ:functional_fgh}), we simply use monotonic increasing piecewise function $f^{(m)}(x)$ instead of continuous monotonic function defined in Eq. (\ref{equ:functional_single}) for approximate solution and the function $f^{(m)}(x)$ is designed as below:
\begin{equation}\label{equ:f_function}
\begin{split}
    f^{(m)}&(x):\ (x^{(m)}_{i_m}, f^{(m)}_{i_m}) \\
    s.t. \ x^{(m)}_1&<x^{(m)}_2<...<x^{(m)}_I, \\
    f^{(m)}_1 &\leq f^{(m)}_2 \leq...\leq f^{(m)}_I, \\
    f^{(m)}_1&=0, f^{(m)}_I=1, \\
    |z^{(m)}_{i_m}|&=length(r^{(m)}_{i_m})=T/I \\
\end{split}
\end{equation}
where $r^{(m)}_{i_m}=\{w^{(m)}_{p_t,q_t}|(x^{(m)}_{i_m}-{\Delta_l}^{(m)}_{i_m})\leq w^{(m)}_{p_t,q_t} < (x^{(m)}_{i_m}+{\Delta_r}^{(m)}_{i_m}),t=1,2,...,T\}$ is the similarity set of the $i_m$-th segment in the $m$-th view, $x^{(m)}_{i_m}=mean(r^{(m)}_{i_m})$, $x^{(m)}_{i_m}+{\Delta_r}^{(m)}_{i_m}=x^{(m)}_{i_m+1}-{\Delta_l}^{(m)}_{i_m+1}$, $m \in [1,2,...,M]$, $i_m \in [1,2,...,I]$, $I$ is the total segments of piecewise function and we divide the data $\{w^{(m)}_{p_t,q_t}\}$ of total $T$ length into $I$ equal parts in the order of $\{w^{(m)}_{p_t,q_t}\}$ from small to large. With the above definitions, we propose three types of functions including single-view function $Fsingle^{(m)}(x):\ (x^{(m)}_{i_m}, Fsingle^{(m)}_{i_m})$, cross-view function $Fcross^{(m)}(x):\ (x^{(m)}_{i_m}, Fcross^{(m)}_{i_m})$ and multi-view function $Fmulti^{(m)}(x):\ (x^{(m)}_{i_m}, Fmulti^{(m)}_{i_m})$, where $i_m \in [1,2,...,I]$. 

\textbf{{\textit{Definition 5:}}} The single-view function is designed as:
\begin{equation}\label{equ:f_single}
\begin{split}
    Fsingle^{(m)}_{i_m}=f^{(m)}_{i_m} = f^{(m)}(x^{(m)}_{i_m}) \\
\end{split}
\end{equation}
where $i_m \in [1,2,..,I]$ and $m \in [1,2,...,M]$. 

\textbf{{\textit{Definition 6:}}} The cross-view function is designed as below to measure the similarity distribution of another cross view ($m$-th view's cross view $b$):
\begin{equation}\label{equ:f_cross}
\begin{split}
    Fcross^{(m)-(b)}_{i_m} &= \frac {1} {|z^{(m)}_{i_m}|} \sum \limits_{x \in r^{(m)}_{i_m},x^{(b)}_{i_b} \in r^{(m)-(b)}_{i_m}} f^{(b)}(x^{(b)}_{i_b}|x) \\
    &= \frac {1} {|z^{(m)}_{i_m}|} \sum \limits_{x \in r^{(m)}_{i_m},x^{(b)}_{i_b} \in r^{(m)-(b)}_{i_m}} f^{(b)}_{i_b} \\
\end{split}
\end{equation}
where $r^{(m)}_{i_m}=\{w^{(m)}_{p_t,q_t}|(x^{(m)}_{i_m}-{\Delta_l}^{(m)}_{i_m})\leq w^{(m)}_{p_t,q_t} < (x^{(m)}_{i_m}+{\Delta_r}^{(m)}_{i_m}),t=1,2,...,T\}$ is the similarity set of the $i_m$-th segment in the $m$-th view, $r^{(m)-(b)}_{i_m}=\{x^{(b)}_{i_x}|(x^{(m)}_{i_m}-{\Delta_l}^{(m)}_{i_m})\leq w^{(m)}_{p_t,q_t} < (x^{(m)}_{i_m}+{\Delta_r}^{(m)}_{i_m}),(x^{(b)}_{i_x}-{\Delta_l}^{(b)}_{i_x})\leq w^{(b)}_{p_t,q_t} < (x^{(b)}_{i_x}+{\Delta_r}^{(b)}_{i_x}),t=1,2,...,T\}$ is the segment set to which the pairwise samples in the $i_m$-th segment in the $m$-th view belongs in the $b$-th view, $|z^{(m)}_{i_m}|=length(r^{(m)}_{i_m})$, $i_m,i_b \in [1,2,...,I]$ and $m,b \in [1,2,...,M]$. 

As designed in Eq. (\ref{equ:p_bayesian_new}), given the pairwise similarity $(x^{(1)}_{i_1},x^{(2)}_{i_2},...,x^{(M)}_{i_M})$ of $M$ views, the joint probability is defined as:
\begin{equation}\label{equ:f_joint}
\begin{split}
    &\ Fjoint(x^{(1)}_{i_1},x^{(2)}_{i_2},...,x^{(M)}_{i_M}) \\
    &= \frac {\prod \limits_{m=1}^M f^{(m)}(x^{(m)}_{i_m})} {\prod \limits_{m=1}^M f^{(m)}(x^{(m)}_{i_m})+\prod \limits_{m=1}^M (1-f^{(m)}(x^{(m)}_{i_m}))} \\
    &= \frac {\prod \limits_{m=1}^M f^{(m)}_{i_m}} {\prod \limits_{m=1}^M f^{(m)}_{i_m}+\prod \limits_{m=1}^M (1-f^{(m)}_{i_m})} \\
    \end{split}
\end{equation}

\textbf{{\textit{Definition 7:}}} The multi-view function is designed as below to measure the similarity distribution of multiple views:
\begin{small}
\begin{equation}\label{equ:f_multi}
\begin{split}
    &Fmulti^{(m)}_{i_m} = \frac {1} {|z^{(m)}_{i_m}|} \sum \limits_{x \in r^{(m)}_{i_m}} Fjoint(x^{(1)}_{i_1},x^{(2)}_{i_2},...,x^{(M)}_{i_M}|x) \\
\end{split}
\end{equation}
\end{small}
where $r^{(m)}_{i_m}=\{w^{(m)}_{p_t,q_t}|(x^{(m)}_{i_m}-{\Delta_l}^{(m)}_{i_m})\leq w^{(m)}_{p_t,q_t} < (x^{(m)}_{i_m}+{\Delta_r}^{(m)}_{i_m}),t=1,2,...,T\}$ is the similarity set of the $i_m$-th segment in the $m$-th view, $x^{(b)}_{i_b} \in r^{(m)-(b)}_{i_m}=\{x^{(b)}_{i_x}|(x^{(m)}_{i_m}-{\Delta_l}^{(m)}_{i_m})\leq w^{(m)}_{p_t,q_t} < (x^{(m)}_{i_m}+{\Delta_r}^{(m)}_{i_m}),(x^{(b)}_{i_x}-{\Delta_l}^{(b)}_{i_x})\leq w^{(b)}_{p_t,q_t} < (x^{(b)}_{i_x}+{\Delta_r}^{(b)}_{i_x}),b \neq m, t=1,2,...,T\}$ is the segment set to which the pairwise samples in the $i_m$-th segment in the $m$-th view belongs in the $b$-th view, $|z^{(m)}_{i_m}|=length(r^{(m)}_{i_m})$, $i_m,i_b \in [1,2,...,I]$ and $m,b \in [1,2,...,M]$.

\subsubsection{{\textbf{Objective Function}}}
\label{sec:sub_function3}
With the above definitions of consistency and multiple probability functions, we propose the following objective function to learn each view’s individual distribution:
\begin{equation}\label{equ:loss_all}
\begin{split}    
    L = \lambda L_{consistency} + L_{constraint}
\end{split}
\end{equation}
where $L_{consistency}$ is consistency loss and $L_{constraint}$ is constraint loss. The parameter $\lambda$ is the balanced factor on $L_{consistency}$ and $L_{constraint}$. 

{\textbf{Consistency Loss.}} The consistency loss aims to learn the consistency from multiple views between $Fsingle^{(m)}_{i_m}$, $Fcross^{(m)-(b)}_{i_m}$ and $Fmulti^{(m)}_{i_m}$. Based on Eq. (\ref{equ:functional_con}), $L_{consistency}$ is defined as:
\begin{small}
\begin{equation}\label{equ:loss_fz}
\begin{split}    
    &L_{consistency} = \frac{1}{M}\sum \limits_{m} (\frac{1}{I}{\sum \limits_{i_m} D(Fsingle^{(m)}_{i_m},Fmulti^{(m)}_{i_m})}) \\
    &+ \frac{1}{M}\sum \limits_{m} (\frac{1}{I} \sum \limits_{i_m} \sum \limits_{b \neq m} D(Fsingle^{(m)}_{i_m},Fcross^{(m)-(b)}_{i_m})) \\
\end{split}
\end{equation}
\end{small}
where $i_m \in [1,2,..,I]$ and $m \in [1,2,...,M]$. Due to the difficulty of consistency loss in Eq. (\ref{equ:loss_fz}) in which $Fsingle$ needs to be constrained by both $Fmulti$ and $Fcross$, the mix function is designed as below for fusion and learning instead of directly learning the consistency between $Fsingle^{(m)}_{i_m}$, $Fcross^{(m)-(b)}_{i_m}$ and $Fmulti^{(m)}_{i_m}$:
\begin{small}
\begin{equation}\label{equ:f_mix}
\begin{split}
    Fmix^{(m)}_{i_m} = \sqrt {Fmulti^{(m)}_{i_m} \frac{1}{M}(Fsingle^{(m)}_{i_m}+\sum \limits_{b \neq m} Fcross^{(m)-(b)}_{i_m})}
\end{split}
\end{equation}
\end{small}
where $i_m \in [1,2,..,I]$ and $m \in [1,2,...,M]$. The mix function $Fmix^{(m)}_{i_m}$ is used to constrain the value of $Fsingle^{(m)}_{i_m}$ and $Fcross^{(m)-(b)}_{i_m}$. Lastly, the consistency loss is mathematically designed as:
\begin{small}
\begin{equation}\label{equ:loss_1}
\begin{split}    
    &L_{consistency1} = \frac{1}{M}\sum \limits_{m} (\frac{1}{I}{\sum \limits_{i_m} D(Fsingle^{(m)}_{i_m},Fmix^{(m)}_{i_m}})) \\
    &L_{consistency2} = \frac{1}{M}\sum \limits_{m} (\frac{1}{I} \sum \limits_{i_m} \sum \limits_{b \neq m} D(Fcross^{(m)-(b)}_{i_m},Fmix^{(m)}_{i_m})) \\
    &L_{consistency} = \frac{1}{M}  (L_{consistency1}+L_{consistency2})
\end{split}
\end{equation}
\end{small}
where $D$ is the distance function and we use $D(x,y)=(x-y)^2$ in our experiments. 
The detailed experiments on consistency loss with Eq. (\ref {equ:loss_fz}) and Eq. (\ref {equ:loss_1}) are listed in Table \ref{tab:loss2}. 

{\textbf{Constraint Loss.}} As defined in Eq. (\ref{equ:f_function}), the value range of the probability function is 0 to 1 and the constraint loss aims to limit the range of the functions, including single-view function $Fsingle^{(m)}(x):\ (x^{(m)}_{i_m}, Fsingle^{(m)}_{i_m})$, cross-view function $Fcross^{(m)}(x):\ (x^{(m)}_{i_m}, Fcross^{(m)}_{i_m})$ and multi-view function $Fmulti^{(m)}(x):\ (x^{(m)}_{i_m}, Fmulti^{(m)}_{i_m})$. Mathematically, the constraint loss is designed as below to limit the values of functions at the beginning and the end:
\begin{small}
\begin{equation}\label{equ:loss_2}
\begin{split}    
    L_{constraint} = \sum \limits_{m} (&\sum \limits_{i_m \in r_i} D(Fmulti^{(m)}_{i_m},0) \\
    &+ \sum \limits_{j_m \in r_j} D(Fmulti^{(m)}_{j_m},1) \\
    &+ \sum \limits_{i_m \in r_i} D(Fsingle^{(m)}_{i_m},0) \\
    &+ \sum \limits_{j_m \in r_j} D(Fsingle^{(m)}_{j_m},1) \\
    &+ \sum \limits_{b \neq m} \sum \limits_{i_m \in r_i} D(Fcross^{(m)-(b)}_{i_m},0) \\
    &+ \sum \limits_{b \neq m} \sum \limits_{j_m \in r_j} D(Fcross^{(m)-(b)}_{j_m},1))
\end{split}
\end{equation}
\end{small}
where $r_i=[1,2,..,indi]$ and $r_j=[I-indj,...,I-1,I]$. $indi$ and $indj+1$ are the limit width and the detailed parameters are listed in Table \ref{tab:setting}. Specially, there is a monotonic constraint in Eq. (\ref{equ:f_function}) which is not included in $L_{constraint}$. For monotonic constraint, we use mandatory constraint to ensure that the functions satisfy monotonicity in the process of iteration.

\subsection{Graph-context-aware Refinement}
\begin{figure}[h!]
    \centering
    \includegraphics[width=0.47\textwidth]{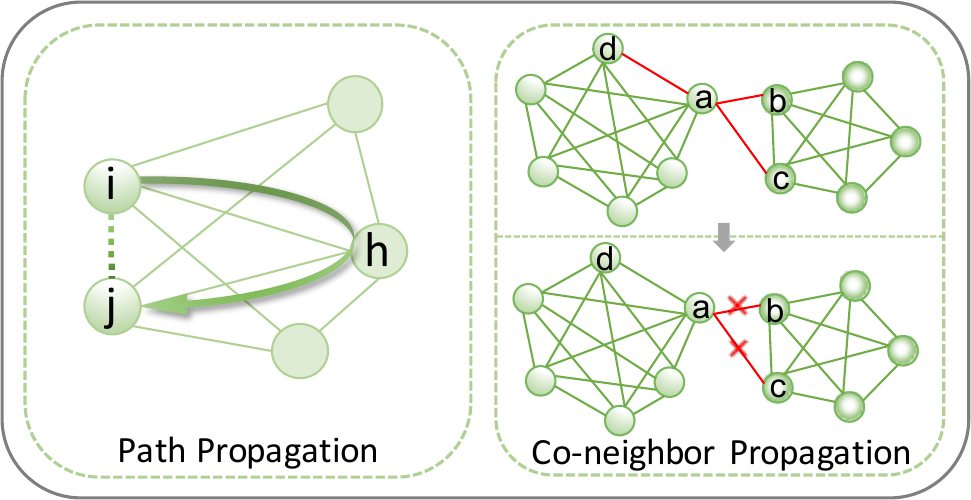}
    \caption{Illustration of the proposed graph-context-aware refinement including path propagation and co-neighbor propagation. As shown in path propagation, taken probability consistency information into consideration, $h$ sets up the probability path between $i$ and $j$ and the probability between $i$ and $j$ can be enhanced by finding the path with the maximum probability. Besides, in co-neighbor propagation, $b$ and $c$ are the noise in k-nearest-neighbors of $a$. Based on the number of common neighbours and the proportion of the common probabilities, co-neighbor propagation refinement adjusts the probability between $a$ and $b$ and the probability between $a$ and $c$ to a small value and the small value indicates that they are not linked. The probability between $a$ and $d$ can be further adjusted and enhanced.}
    \label{fig:refine}
\end{figure}
The probability estimation in Eq. (\ref{equ:p_bayesian_new}) is calculated based on the aspect of sample relationship, overlooking the aspect of graph context which contains rich information. Thus, we perform graph-context-aware refinement with path propagation and co-neighbor propagation to further alleviate the impact of noise and outliers.

Due to the data perturbation of each view, there exists a few outliers in dataset which may affect the clustering performance in the final step. The probability estimation of outliers can not be calculated accurately by using Eq. (\ref{equ:p_bayesian_new}), we therefore try to fine-tune them with path propagation. Inspired by the message passing, where the information among nodes is transmissible, the proposed path propagation passes probabilities between samples like follows:
\begin{equation}\label{equ:pb_optpath}
    P(i,j) = \max {(P(i,j), P(i,h) \times P(h,j))}
\end{equation}
where $j\in knn_i$, $h\in knn_{ij}$, $knn_i=\{\cup knn^m_i\}$, $knn_j=\{\cup knn^m_j\}$, $knn_{ij}=\{{knn_i \cap knn_j}\}$ and $knn^m_i\in R^{k}$ is the k-nearest-neighbors of sample $i$ in $m$-th view. Fig. \ref{fig:refine} shows an intuitive path propagation case, in which sample $h$ sets up the path between sample $i$ and sample $j$ and the probability between sample $i$ and sample $j$ can be enhanced by finding the path with the maximum probability. From the aspect of probability, given three samples (sample $i$, $j$, $h$) and let $a=P(i,j)$, $b=P(i,h), c=P(j,h)$ for short, the probability that sample $i$ and sample $j$ belong to one class is defined as $q=q_p / q_a$, where $q_p=abc+a(1-b)(1-c)$, $q_a=abc+a(1-b)(1-c)+(1-a)(1-b)(1-c)+(1-a)(1-b)c+(1-a)b(1-c)$. In the formula, $q_a$ denotes the sum of all possibilities and $q_p$ denotes the sum of all possibilities that sample $i$ and sample $j$ belong to one class. Simply given $P(i,j)=0.5$ for a fuzzy probability, it's natural to prove:
\begin{equation}\label{equ:pb_optgps_max}
    q=\frac{q_p}{q_a}=\frac{bc+\frac{1}{2}(1-b-c)}{\frac{1}{2}bc+1-\frac{1}{2}b-\frac{1}{2}c} \geq bc 
\end{equation}
where $0 < b,c < 1$. Using path propagation, the probability consistency information between the outliers and their neighbors is taken into consideration, in which the outliers can be detected and the pairwise probabilities between the outliers and their neighbors can be enhanced.

Besides, the probability estimation is calculated in Euclidean space while the visual features usually lie in low-dimensional manifolds\cite{Manifolds}. Only using the information in Euclidean space, overlooking the graph context, may result in inaccuracy of the actual pairwise posterior probabilities between samples. To take advantage of the graph context, the co-neighbor propagation is defined as:
\begin{equation}\label{equ:pb_optknn}
    P(i,j) = \frac{\sum_{h\in knn_{ij}} (P(i,h)+P(j,h))} {\sum_{h_i\in knn_i} {P(i,h_i)} + \sum_{h_j\in knn_j} {P(j,h_j)}}
\end{equation}
where $knn_i\in R^{k}$ is the k-nearest-neighbors of sample $i$ calculated by $P(i,j)$ and $knn_{ij}=\{{knn_i \cap knn_j}\}$.  Fig. \ref{fig:refine} shows an intuitive co-neighbor propagation case, in which the local graph is constructed by the k-nearest-neighbors of two samples. We take both the number of common neighbours and the proportion of the common probabilities into consideration to further refine the probability based on the local graph. As shown in Eq. (\ref{equ:pb_optknn}), the available graph-based probability information can be mined to dig out as much manifold-like distribution information as possible. Using co-neighbor propagation, the noise in k-nearest-neighbors can be detected and the outliers can be further enhanced in an efficient way.

\subsection{Probabilistic Clustering}
\begin{figure}[h!]
    \centering
    \includegraphics[width=0.47\textwidth]{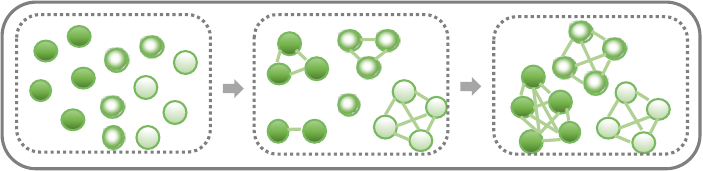}
    \caption{Illustration of the proposed probabilistic clustering. Each sample is assigned to its own clustering set at the beginning and each sample is moved to the neighbour clustering set in random sequential order by maximizing joint probability iteratively. Finally, a good clustering result can be generated in a convergent way.}
    \label{fig:probclustering}
\end{figure}
Given the estimated self-learning probability function, we can utilize Eq. (\ref{equ:p_bayesian_new}) to calculate the multi-view pairwise posterior matching probability $P(i,j)$ and we can utilize graph-context-aware refinement to further refine the probability $P(i,j)$. Finally, to cluster samples in an unsupervised manner, the probabilistic clustering algorithm is introduced to generate clustering result without any prior knowledge based on the probability $P(i,j)$. Given $N$ samples with the clustering set $\pi:[z_1,z_2,...,z_N]$, the optimization goal of probabilistic clustering can be mathematically expressed as:
\begin{equation}\label{equ:pi_opt}
\begin{split}
    \pi_{opt}=\mathop{\mathrm{argmax}}\limits_{\pi}&{P(X|\pi)}=\mathop{\mathrm{argmax}}\limits_{\pi}{\frac{P(X,\pi)}{P(\pi)}} \\
    s.t.\ P(X,\pi)=\ &\frac{\prod_{i,j} (\frac{P(e_{ij}=1)}{P(e_{ij}=0)})^{\delta (z_i,z_j)}P(e_{ij}=0)}{\Omega}\\
\end{split}
\end{equation}
where $\delta$ is the Kronecker function and $\Omega$ is the normalization parameter. Besides, there exists an easy-to-understand formula for probabilistic clustering and $P(X,\pi)$ can be mathematically expressed as:
\begin{equation}\label{equ:pi_opt_origin}
\begin{split}
    P(X,\pi)=\ &\frac{\prod_{i,j} P(e_{ij}=1)^{\delta (z_i,z_j)}P(e_{ij}=0)^{1-\delta (z_i,z_j)}}{\Omega}\\
\end{split}
\end{equation}
The basic idea of probabilistic clustering is to maximize the intra-cluster similarities and minimize the inter-cluster similarities among the samples and Eq. (\ref{equ:pi_opt}) and Eq. (\ref{equ:pi_opt_origin}) are equivalent. With the above definitions, the objective optimization function $L = -logP(X|\pi)$ can be expressed as:
\begin{equation}\label{equ:pi_loss}
    L=\sum_{i,j} (\delta (z_i,z_j)(logP(e_{ij}=0)-logP(e_{ij}=1))) + c
\end{equation}
where $c = -\sum_{i,j} (logP(e_{ij}=0)) - logP(\pi) - log{\Omega}$ is a constant. Only the probabilities within the class need to be calculated in Eq. (\ref{equ:pi_loss}), which reduces the computational complexity. The whole probabilistic clustering optimization procedure is outlined in Algorithm \ref{alg:MVC} and Fig. \ref{fig:probclustering} shows an intuitive clustering process. In the first step, k-nearest-neighbors is constructed using refined multi-view pairwise posterior matching probability. In the second step, each sample is assigned to its own clustering set. Then, in random sequential order, each sample is moved to the neighbour clustering set that results in the minimum value using Eq. (\ref{equ:pi_loss}). The moving procedure is repeated for every sample until no moving steps. With this algorithm, a good clustering result can be generated in a convergent way.

\begin{algorithm}
    \caption{Probabilistic Clustering Optimization Procedure}
    \label{alg:MVC}
    Input: $P(e_{ij}=1)$ and $P(e_{ij}=0)$\;
    Construct KNN $nbrs\in R^{n*k}$ by $P(e_{ij}=1)$\;
    Initialization: $listn=[1,2,...,n]$, $it=0$, $maxiter=20$, $z=[z_1,z_2,...,z_n]=[1,2,..,n]$\;
    \While{$it < maxiter$}{
        $count=0$\\
        random shuffle $listn$\\
        \For{$i$ in $listn$}{
            find $z_{find}$ in $z[nbrs[i]]$ with minimum objective value denoted by Eq. (\ref{equ:pi_loss}) \\
            \If{$z_i\ != z_{find}$}{
                update $z_i = z_{find}$\\
                $count=count+1$\\
            }
        }
        \If{$count==0$}{
            break\\
        }
        $it=it+1$\\
    }
    Output: $z$\;
\end{algorithm}

\begin{algorithm}
    \caption{Summary of SLS-MPC}
    \label{alg:sum}
    Input: a multi-view dataset of $N$ samples with $M$ views $S=\{V^{(1)},V^{(2)},...,V^{(M)}\}$\;
    Solution: \ \\
    \ \ \ \ \ 1.\ Construct $KNN^{(m)} \in R^{N*K}$ based on the similarity matrix $W^{(m)} \in R^{N*N}$ of the $m$-th view; Construct the training data including total $T$ pairwise samples $(p_t,q_t)$ and the corresponding similarity values $(w^{(1)}_{p_t,q_t},w^{(2)}_{p_t,q_t},...,w^{(M)}_{p_t,q_t})$\;
    \ \ \ \ \ 2.\ Using Eq. (\ref{equ:loss_all}), Eq. (\ref{equ:loss_1}) and Eq. (\ref{equ:loss_2}) to learn probability function $P(e_{ij}=1|w^{(m)}_{ij})$\;
    \ \ \ \ \ 3.\ Using Eq. (\ref{equ:p_bayesian_new}) to estimate the pairwise posterior probability $P(e_{ij}=0/1)$ of sample $i$ and $j$\;
    \ \ \ \ \ 4.\ Using Eq. (\ref{equ:pb_optpath}) and Eq. (\ref{equ:pb_optknn}) to further refine the pairwise probability $P(e_{ij}=0/1)$\;
    \ \ \ \ \ 5.\ Using Algorithm \ref{alg:MVC} to perform probabilistic clustering based on the refined pairwise probability $P(e_{ij}=0/1)$ and generate clustering results $z$\;
    Output: $z$\;
\end{algorithm}

\subsection{Summary of SLS-MPC}
In this section, we summarize the whole framework of SLS-MPC. Firstly, SLS-MPC proposes a self-learning probability function to learn $P(e_{ij}=1|w^{(m)}_{ij})$ using Eq. (\ref{equ:loss_all}), Eq. (\ref{equ:loss_1}) and Eq. (\ref{equ:loss_2}). Then the pairwise posterior probability $P(e_{ij}=0/1)$ of sample $i$ and $j$ is estimated using the proposed symmetric multi-view probability estimation formula in Eq. (\ref{equ:p_bayesian_new}). Next SLS-MPC uses Eq. (\ref{equ:pb_optpath}) and Eq. (\ref{equ:pb_optknn}) to further refine the pairwise probability based on the the aspect of graph context. Finally, the refined pairwise probability $P(e_{ij}=0/1)$ is used as input to the probabilistic clustering optimization procedure to generate clustering results.

\section{Experiments}
\label{sec:experiments}
\subsection{Experimental Settings}
\begin{table}[htbp]
  \centering
  \caption{Summary of the datasets. \{$M$, $C$, $N$, $d^{(m)}$\} denotes the number of \{views, clusters, samples, features\} in each view, respectively.}
    \begin{tabular}{|c|cccc|}
    \hline
    Datasets & $M$ & $C$ & $N$ & $d^{(m)}(m=1,...,M)$ \bigstrut\\
    \hline
    Handwritten & 4     & 10    & 2000  & 240,76,47,64 \bigstrut[t]\\
    100Leaves & 2     & 100   & 1600  & 64,64 \\
    Humbi240 & 2     & 240   & 13440 & 256,256 \\
    BUAA  & 2     & 150   & 1350  & 100,100 \\
    BBCSport & 2     & 5     & 544   & 3181,3202 \bigstrut[b]\\
    \hline
    \end{tabular}%
  \label{tab:datasets}%
\end{table}%

\begin{table}[htbp]
  \centering
  \caption{The detailed settings of $I,indi,indj$ and $\lambda$.}
    \begin{tabular}{|c|cccc|}
    \hline
    Datasets & $I$     & $indi$  & $indj+1$  & $\lambda$ \bigstrut\\
    \hline
    Handwritten view1-4 & 1000  & 10    & 4     & 80 \bigstrut[t]\\
    Handwritten view1-2 & 1000  & 10    & 4     & 20 \\
    100Leaves & 200   & 10    & 2     & 2 \\
    Humbi240 & 1000  & 10    & 4     & 20 \\
    BUAA  & 200   & 10    & 4     & 20 \\
    BBCSport & 200   & 10    & 4     & 20 \bigstrut[b]\\
    \hline
    \end{tabular}%
  \label{tab:setting}%
\end{table}%

\noindent{\bf Datasets.} The experimental comparisons are experimentally evaluated on several multi-view datasets. {\bf (1) Handwritten}\cite{HW} contains 2000 samples of 10 digits (i.e., digits '0-9'), covering four kinds of features, which are average pixels features, Fourier coefficient features, Zernike moments features and Karhunen-Love coefficient features. {\bf (2) 100Leaves}\cite{100Leaves} contains 1600 samples from 100 plant species. For each sample, a shape descriptor and texture histogram are given. {\bf (3) Humbi240}, a subset of Humbi\cite{HUMBI} dataset, contains 13440 samples of 240 persons covering face features extracted by face recognition model\footnote{https://github.com/XiaohangZhan/face\_recognition\_framework} and body features extracted by person reID model\footnote{https://github.com/layumi/Person\_reID\_baseline\_pytorch}. {\bf (4) BUAA-visnir face dataset (BUAA)}\cite{BUAA} contains 1350 visual images and 1350 near infrared images of the 150 volunteers. {\bf (5) BBCSport}\footnote{http://mlg.ucd.ie/datasets/segment.html} contains 544 samples of 5 categories. The feature dimensions of the two views used in experiments are 3181 and 3202 respectively. The datasets are summarized in Table \ref{tab:datasets}. To evaluate the clustering performance on incomplete data, we select $c\%$ ($c=90,70,50,30$) samples as the paired samples that have full views. For the remaining samples, half of them miss the first view, while the second view of the other half is removed. The missing rate is defined as $\eta=1-c$.

\noindent{\bf Evaluation Metrics.} In the experiments, several widely-used clustering metrics including BCubed Fmeasure, Pairwise Fmeasure\cite{EM}, Normalized Mutual Information (NMI) and Adjusted Rand Index (ARI) are used as the evaluation metrics. A higher value of these metrics indicates a better clustering performance.

\noindent{\bf Implementation Details.} We implement our SLS-MPC in PyTorch 1.2\cite{PYTORCH} and perform all evaluations on a standard Linux OS with 16 2.50GHz Intel Xeon Platinum 8163 CPUs. The self-learning probability function of each view is initialized as a uniform line from 0 to 1 and the self-learning probability function is trained by SGD with a learning rate of 0.001, a momentum of 0.9 and a weight decay of 0.00005. The detailed settings of $I,indi,indj$ and $\lambda$ are listed in Table \ref {tab:setting}. The setting of $I$ takes into account the size of training data $T$. 

\begin{table*}[htbp]
  \centering
  \caption{The clustering performance comparisons on three datasets. MVC indicates complete multi-view clustering; IMVC indicates incomplete multi-view clustering with 0.5 missing rate.}
    \begin{tabular}{|c|c|cccc|cccccccc|}
    \hline
    \multirow{2}[2]{*}{Type} & \multirow{2}[2]{*}{Methods} & \multicolumn{4}{c|}{Handwritten} & \multicolumn{4}{c|}{100Leaves} & \multicolumn{4}{c|}{Humbi240} \bigstrut[t]\\
          &       & $F_P$ & $F_B$    & NMI   & ARI   & $F_P$ & $F_B$    & NMI   & \multicolumn{1}{c|}{ARI} & $F_P$ & $F_B$    & NMI   & ARI \bigstrut[b]\\
    \hline
    \multirow{11}[2]{*}{MVC} & MCDCF\cite{MCDCF} & 54.92  & 59.32  & 64.90  & 49.45  & 51.04  & 58.14  & 82.20  & \multicolumn{1}{c|}{50.52 } & 53.16  & 67.99  & 88.91  & 52.91  \bigstrut[t]\\
          & SMSC\cite{SMSC}  & 67.48  & 69.20  & 72.54  & 63.83  & 25.88  & 42.12  & 72.59  & \multicolumn{1}{c|}{24.77 } & 26.59  & 44.37  & 74.09  & 26.13  \\
          & SFMC\cite{SFMC}  & 72.70  & 73.72  & 77.35  & 69.66  & 29.97  & 61.31  & 80.97  & \multicolumn{1}{c|}{28.94 } & 51.78  & 91.19  & 95.47  & 51.50  \\
          & IMCCP\cite{IMCCP} & 76.56  & 80.96  & 83.86  & 73.73  & 22.91  & 36.20  & 69.94  & \multicolumn{1}{c|}{21.78 } & 49.68  & 58.43  & 88.42  & 49.37  \\
          & GMC\cite{GMC}   & 74.84  & 80.47  & 82.20  & 71.75  & 36.40  & 78.98  & 88.75  & \multicolumn{1}{c|}{35.47 } & 87.99  & 96.05  & 98.57  & 87.94  \\
          & OSLF\cite{OSLF}  & 78.24  & 78.55  & 79.32  & 75.82  & 65.55  & 69.59  & 87.68  & \multicolumn{1}{c|}{65.20 } & 90.35  & 93.62  & 98.20  & 90.31  \\
          & EEIMC\cite{EEIMC} & 78.86  & 79.13  & 80.80  & 76.51  & 74.10  & 77.53  & 91.18  & \multicolumn{1}{c|}{73.84 } & 91.45  & 94.45  & 98.54  & 91.41  \\
          & UEAF\cite{UEAF}  & 80.61  & 80.92  & 81.43  & 78.46  & 64.54  & 72.81  & 89.18  & \multicolumn{1}{c|}{64.16 } & 86.36  & 90.36  & 97.11  & 86.30  \\
          & PIC\cite{PIC}   & 76.61  & 77.88  & 80.23  & 73.94  & 78.04  & 81.49  & 92.76  & \multicolumn{1}{c|}{77.82 } & 94.34  & 96.29  & 98.95  & 94.32  \\
          & MPC\cite{MPC}   & 84.57  & 84.45  & 85.60  & 83.04  & 84.18  & 85.65  & 94.40  & \multicolumn{1}{c|}{84.04 } & 95.49  & 97.03  & 99.07  & 95.47  \\
          & SLS-MPC  & \textbf{87.03 } & \textbf{86.51 } & \textbf{87.62 } & \textbf{85.73 } & \textbf{85.46 } & \textbf{86.39 } & \textbf{95.03 } & \multicolumn{1}{c|}{\textbf{85.34 }} & \textbf{98.12 } & \textbf{98.77 } & \textbf{99.62 } & \textbf{98.11 } \bigstrut[b]\\
    \hline
    \multirow{11}[2]{*}{IMVC} & MCDCF\cite{MCDCF} & 20.84  & 22.99  & 25.38  & 11.38  & 23.84  & 30.61  & 68.36  & \multicolumn{1}{c|}{23.06 } & 29.91  & 41.78  & 71.44  & 29.53  \bigstrut[t]\\
          & SMSC\cite{SMSC}  & 62.83  & 63.26  & 65.65  & 58.65  & 17.51  & 30.59  & 63.26  & \multicolumn{1}{c|}{16.27 } & 18.69  & 31.59  & 64.42  & 18.17  \\
          & SFMC\cite{SFMC}  & 54.81  & 67.30  & 71.99  & 47.53  & 22.67  & 51.94  & 73.81  & \multicolumn{1}{c|}{21.50 } & 7.61  & 71.73  & 81.66  & 6.88  \\
          & IMCCP\cite{IMCCP} & 58.52  & 71.10  & 72.68  & 52.71  & 17.08  & 24.75  & 60.84  & \multicolumn{1}{c|}{15.99 } & 37.20  & 42.66  & 80.93  & 36.84  \\
          & GMC\cite{GMC}   & 53.56  & 73.19  & 73.56  & 46.05  & 3.55  & 47.35  & 56.76  & \multicolumn{1}{c|}{1.76 } & 2.55  & 52.86  & 65.28  & 1.75  \\
          & OSLF\cite{OSLF}  & 53.86  & 54.06  & 58.51  & 48.73  & 33.86  & 39.04  & 71.84  & \multicolumn{1}{c|}{33.19 } & 70.72  & 73.40  & 89.41  & 70.59  \\
          & EEIMC\cite{EEIMC} & 68.80  & 69.48  & 70.26  & 65.33  & 52.65  & 56.74  & 81.11  & \multicolumn{1}{c|}{52.18 } & 80.94  & 86.24  & 94.84  & 80.86  \\
          & UEAF\cite{UEAF}  & 68.94  & 69.48  & 72.55  & 65.48  & 38.47  & 45.87  & 75.62  & \multicolumn{1}{c|}{37.82 } & 86.04  & 89.96  & 96.81  & 85.98  \\
          & PIC\cite{PIC}   & 75.65  & 76.03  & 76.67  & 72.95  & 50.79  & 55.61  & 80.72  & \multicolumn{1}{c|}{50.30 } & 83.30  & 85.74  & 94.64  & 83.23  \\
          & MPC\cite{MPC}   & 77.44  & 77.65  & 78.52  & 75.13  & 58.31  & 61.19  & 83.39  & \multicolumn{1}{c|}{57.94 } & 90.10  & 91.56  & 96.53  & 90.06  \\
          & SLS-MPC  & \textbf{77.80 } & \textbf{78.65 } & \textbf{79.62 } & \textbf{75.46 } & \textbf{59.91 } & \textbf{62.87 } & \textbf{84.16 } & \textbf{59.56 } & \textbf{92.69 } & \textbf{94.02 } & \textbf{97.55 } & \textbf{92.66 } \bigstrut[b]\\
    \hline
    \end{tabular}%
  \label{tab:res-3datasets}%
\end{table*}%

\begin{figure*}
    \centering
    \includegraphics[width=0.31\textwidth]{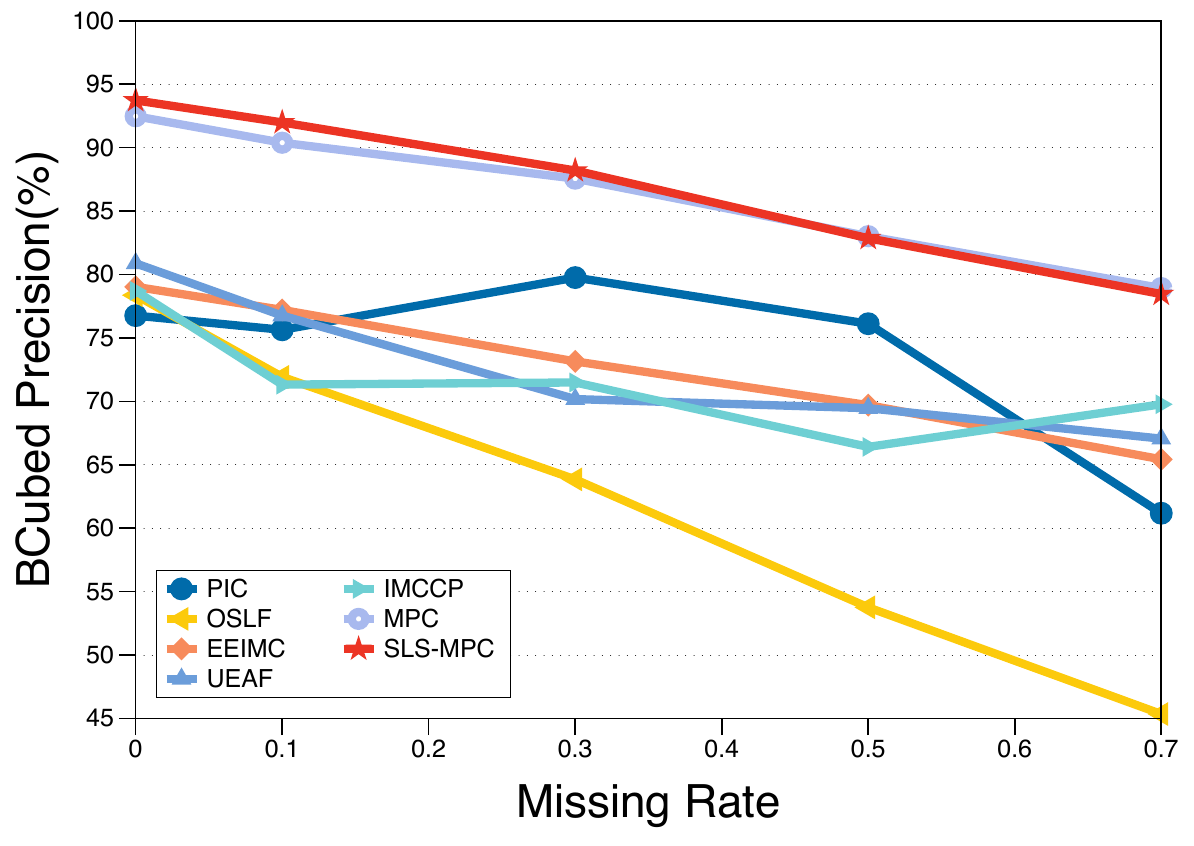}
    \includegraphics[width=0.31\textwidth]{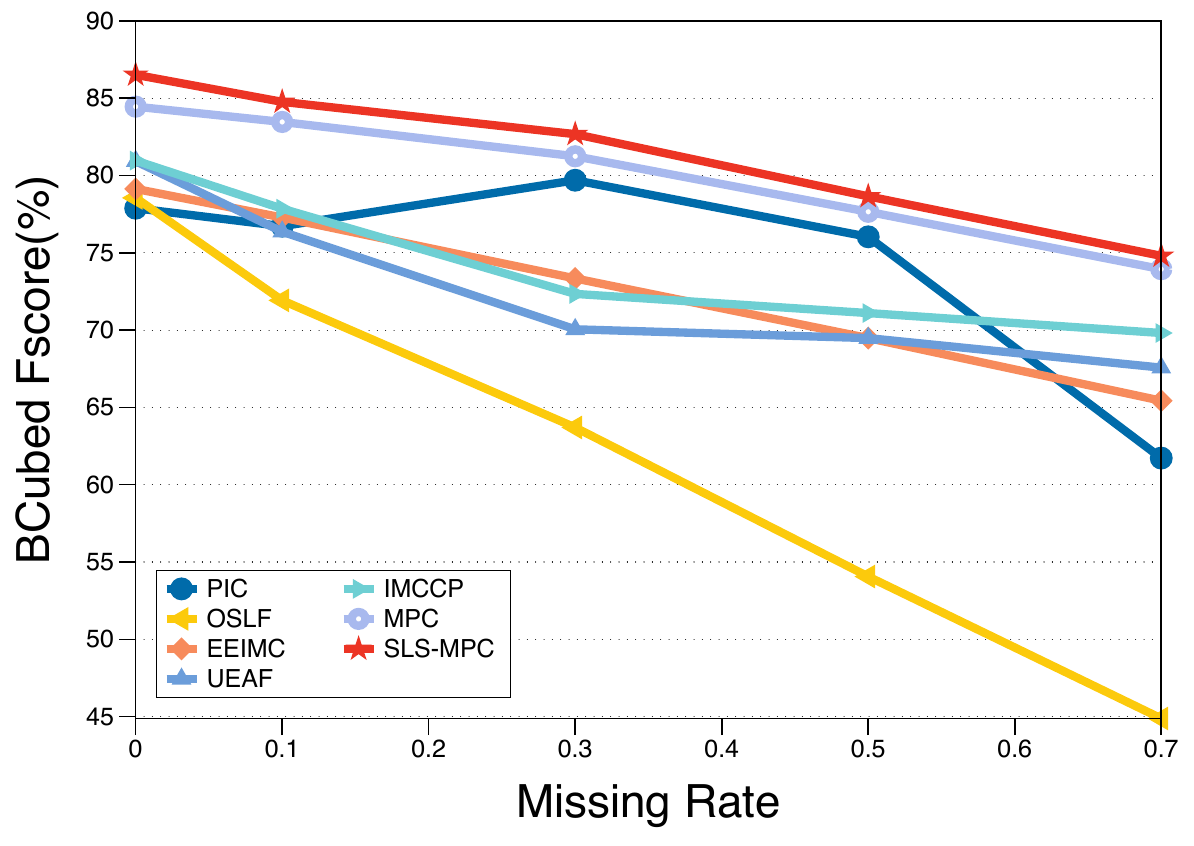}
    \includegraphics[width=0.31\textwidth]{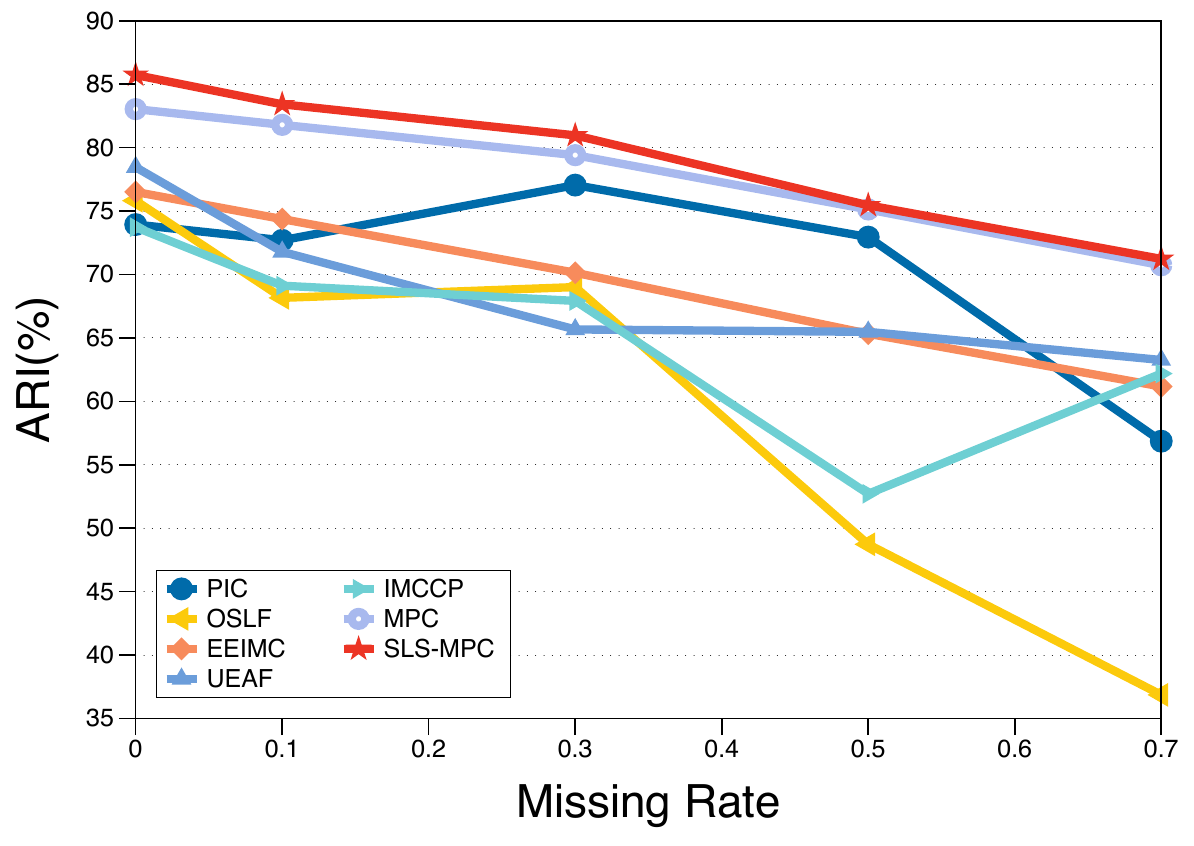}  
    \includegraphics[width=0.31\textwidth]{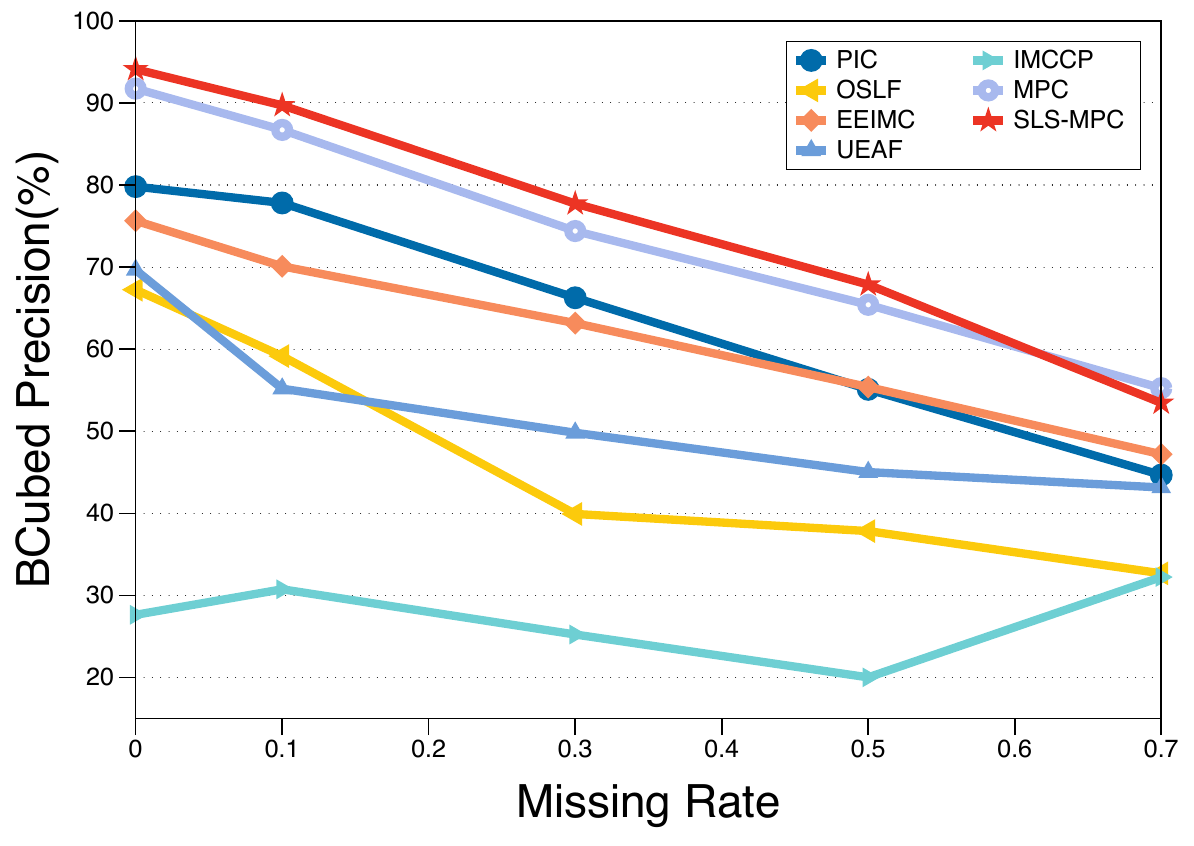}
    \includegraphics[width=0.31\textwidth]{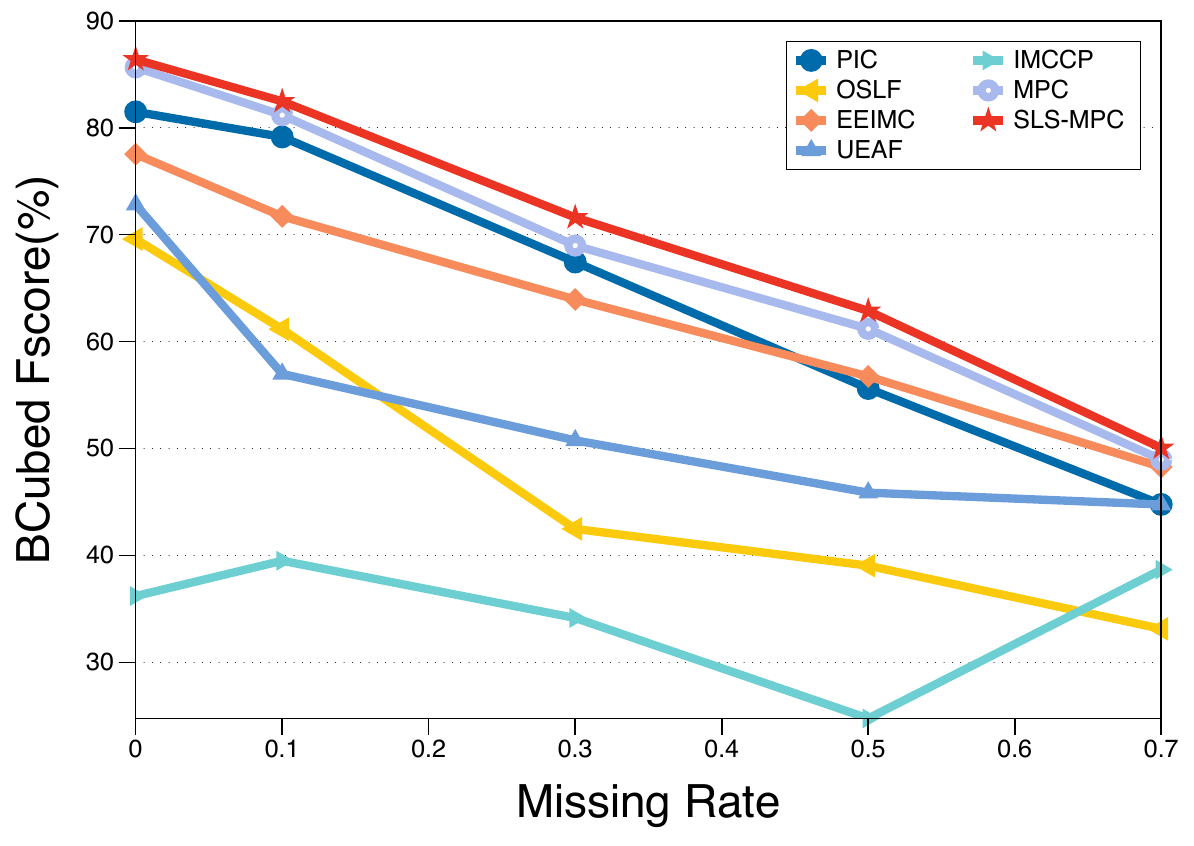}
    \includegraphics[width=0.31\textwidth]{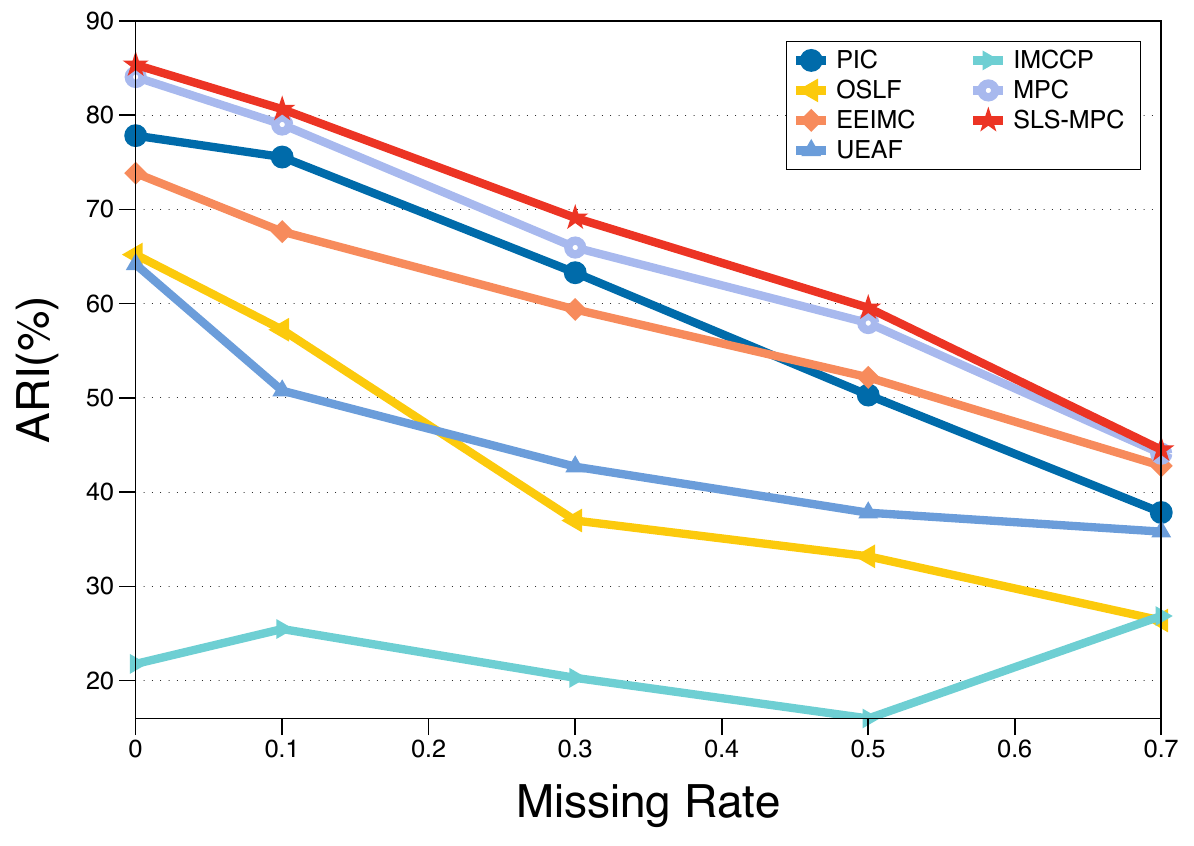}
    \caption{The clustering performance comparisons on Handwritten and 100Leaves with different missing rates. Three comparisons in the first row are experiments on Handwritten. Three comparisons in the second row are experiments on 100Leaves.}
    \label{fig:miss_diff}
\end{figure*}

\subsection{Compared Methods}
We compare our method with SOTA multi-view clustering algorithms. SMSC\cite{SMSC}, GMC\cite{GMC}, MCDCF\cite{MCDCF} and SFMC\cite{SFMC} could only handle complete multi-view data and thus we fill the missing data with the mean values of the same view following previous work\cite{IMCCP} for incomplete clustering cases. PIC\cite{PIC}, OSLF\cite{OSLF}, EEIMC\cite{EEIMC}, UEAF\cite{UEAF}, IMCCP\cite{IMCCP} and MPC\cite{MPC} are six compared methods for complete and incomplete clustering cases. For all methods, we download their released codes and tune the hyper-parameters by grid search to generate the best possible results on each dataset.

\begin{table*}[htbp]
  \centering
  \caption{The clustering performance comparisons on Handwritten with 4 views. View 1 and view 2 are complete and view 3 and view 4 are 50\% missing in the incomplete cases.}
    \begin{tabular}{|c|c|cccccccc|}
    \hline
    \multirow{2}[2]{*}{Type} & \multirow{2}[2]{*}{Methods} & \multicolumn{3}{c}{Pairwise Fmeasure} & \multicolumn{3}{c}{BCubed Fmeasure} & \multirow{2}[2]{*}{NMI} & \multirow{2}[2]{*}{ARI} \bigstrut[t]\\
          &       & Precision & Recall & \multicolumn{1}{p{5em}}{Fscore} & Precision & Recall & \multicolumn{1}{p{5em}}{Fscore} &       &  \bigstrut[b]\\
    \hline
    \multirow{7}[2]{*}{MVC} & OSLF\cite{OSLF}  & 76.23  & 76.58  & 76.40  & 76.28  & 76.70  & 76.49  & 76.51  & 73.79  \bigstrut[t]\\
          & EEIMC\cite{EEIMC} & 75.33  & 76.39  & 75.86  & 76.53  & 76.51  & 76.52  & 78.28  & 73.17  \\
          & PIC\cite{PIC}   & 80.76  & 80.91  & 80.84  & 81.28  & 81.01  & 81.14  & 83.26  & 78.72  \\
          & UEAF\cite{UEAF}  & 81.59  & 82.25  & 81.92  & 82.57  & 82.34  & 82.45  & 83.00  & 79.91  \\
          & IMCCP\cite{IMCCP} & -     & -     & -     & -     & -     & -     & -     & - \\
          & MPC\cite{MPC}   & 95.85  & 85.12  & 90.17  & 94.89  & 85.19  & 89.78  & 89.77  & 89.15  \\
          & SLS-MPC  & \textbf{96.51 } & \textbf{90.25 } & \textbf{93.28 } & \textbf{95.85 } & \textbf{90.30 } & \textbf{92.99 } & \textbf{92.13 } & \textbf{92.56 } \bigstrut[b]\\
    \hline
    \multirow{7}[2]{*}{IMVC} & OSLF\cite{OSLF}  & 62.25  & 67.05  & 64.56  & 64.61  & 67.21  & 65.88  & 69.75  & 60.48  \bigstrut[t]\\
          & EEIMC\cite{EEIMC} & 73.93  & 78.60  & 78.26  & 78.88  & 78.71  & 78.79  & 79.53  & 75.85  \\
          & PIC\cite{PIC}   & 77.24  & 79.72  & 78.46  & 78.83  & 79.82  & 79.32  & 81.34  & 76.04  \\
          & UEAF\cite{UEAF}  & 81.31  & 81.77  & 81.54  & 81.90  & 81.86  & 81.88  & 82.39  & 79.49  \\
          & IMCCP\cite{IMCCP} & -     & -     & -     & -     & -     & -     & -     & - \\
          & MPC\cite{MPC}   & 95.42  & 83.84  & 89.26  & 94.09  & 83.93  & 88.72  & 88.70  & 88.16  \\
          & SLS-MPC  & \textbf{96.77 } & \textbf{87.18 } & \textbf{91.73 } & \textbf{96.00 } & \textbf{87.25 } & \textbf{91.42 } & \textbf{90.92 } & \textbf{90.86 } \bigstrut[b]\\
    \hline
    \end{tabular}%
  \label{tab:res-4v}%
\end{table*}%

\noindent{\bf Performance Comparison with Two Views.} Table \ref{tab:res-3datasets} lists the experimental results of different methods on Handwritten, 100Leaves and Humbi240. In the complete cases, our proposed SLS-MPC achieves the best performance and surpasses the best baseline by 2.69\% on Handwritten, 1.30\% on 100Leaves and 2.64\% on Humbi240 in terms of ARI. Moreover, in the incomplete cases, SLS-MPC surpasses the SOTA by 0.33\% on Handwritten, 1.62\% on 100Leaves and 2.60\% on Humbi240 in terms of ARI. Table \ref{tab:res-add} lists the experimental results of different methods on BUAA and BBCSport and our method surpasses almost all tested baselines in terms of BCubed Precision and Fscore. Furthermore, the incomplete multi-view clustering performance with different missing rates on Handwritten and 100Leaves are shown in Fig. \ref{fig:miss_diff}. From these experimental results, we can observe the following points: (1) our proposed SLS-MPC outperforms all the tested baselines with different missing rates, which demonstrates SLS-MPC's adaptability to different missing rates; (2) SLS-MPC achieves the best precision with almost different missing rates, which further proves the accuracy of self-learning probability function and symmetric multi-view probability estimation in our proposed method.

\begin{table}[htbp]
  \centering
  \caption{The clustering performance of BCubed Precision $Pre_B$ and Fscore $F_B$ comparisons on BUAA and BBCSport. MVC indicates complete multi-view clustering; IMVC indicates incomplete multi-view clustering with 0.5 missing rate.}
    \begin{tabular}{|c|c|cc|cc|}
    \hline
    \multirow{2}[2]{*}{Type} & \multirow{2}[2]{*}{Methods} & \multicolumn{2}{c|}{BUAA} & \multicolumn{2}{c|}{BBCSport} \bigstrut[t]\\
          &       & Precision & Fscore & Precision & Fscore \bigstrut[b]\\
    \hline
    \multirow{7}[2]{*}{MVC} & IMCCP\cite{IMCCP} & 39.29  & 39.74  & 28.67  & 35.42  \bigstrut[t]\\
          & OSLF\cite{OSLF}   & 23.39  & 24.75  & 86.04  & 86.01  \\
          & EEIMC\cite{EEIMC} & 34.09  & 34.49  & 76.87  & 73.71  \\
          & UEAF\cite{UEAF}  & 28.46  & 29.59  & 82.69  & 83.88  \\
          & PIC\cite{PIC}   & 44.25  & 43.65  & 90.41  & 90.39  \\
          & MPC\cite{MPC}   & 58.36  & 44.52  & \textbf{95.52 } & 93.84  \\
          & SLS-MPC  & \textbf{79.22 } & \textbf{49.50 } & 95.04  & \textbf{94.68 } \bigstrut[b]\\
    \hline
    \multirow{7}[2]{*}{IMVC} & IMCCP\cite{IMCCP} & 32.50  & 32.94  & 25.13  & 34.20  \bigstrut[t]\\
          & OSLF\cite{OSLF}   & 30.55  & 31.08  & 66.00  & 63.75  \\
          & EEIMC\cite{EEIMC} & 32.33  & 32.73  & 76.63  & 74.88  \\
          & UEAF\cite{UEAF}  & 29.02  & 30.05  & 87.51  & 87.20  \\
          & PIC\cite{PIC}   & 35.02  & 35.46  & 86.80  & 86.96  \\
          & MPC\cite{MPC}   & 40.56  & 36.84  & 88.45  & 88.34  \\
          & SLS-MPC  & \textbf{44.88 } & \textbf{39.25 } & \textbf{91.01 } & \textbf{90.44 } \bigstrut[b]\\
    \hline
    \end{tabular}%
  \label{tab:res-add}%
\end{table}%

\noindent{\bf Performance Comparison with Four Views.} For the Handwritten dataset, additional incomplete case is constructed in which all samples have two complete views (the first view and the second view) and half of them miss the third view, while the other half of the samples remove the fourth view. As shown in Table \ref{tab:res-4v}, SLS-MPC significantly outperforms these state-of-the-art methods and SLS-MPC surpasses the best baseline by 3.41\% and 2.70\% in terms of ARI in complete case and incomplete case, respectively. The encouraging performance demonstrates SLS-MPC's capacity of extending to multiple views and self-learning capacity of probability function in multi-view information excavation. IMCCP can only handle two views, so the result of IMCCP is not listed in Table \ref{tab:res-4v}. Specially in this case, view completion is introduced to handle data missing $Fjoint(x^{(1)}_{i_1},x^{(2)}_{i_2})$ with only two views, $Fjoint(x^{(1)}_{i_1},x^{(2)}_{i_2},x^{(3)}_{i_3})$ and $Fjoint(x^{(1)}_{i_1},x^{(2)}_{i_2},x^{(4)}_{i_4})$ with only three views. The pairwise probability is defined as in this case:
\begin{small}
\begin{equation}\label{equ:f_joint_com1}
\begin{split}
    &\ Fjoint(x^{(1)}_{i_1},x^{(2)}_{i_2},x^{(3)}_{i_3}) \\
    &= \frac {f^{(1)}_{i_1}f^{(2)}_{i_2}f^{(3)}_{i_3}f^{(4)}_{c}} {f^{(1)}_{i_1}f^{(2)}_{i_2}f^{(3)}_{i_3}f^{(4)}_{c}+(1-f^{(1)}_{i_1})(1-f^{(2)}_{i_2})(1-f^{(3)}_{i_3})(1-f^{(4)}_{c})} \\
    \end{split}
\end{equation}
\end{small}
\begin{small}
\begin{equation}\label{equ:f_joint_com2}
\begin{split}
    &\ Fjoint(x^{(1)}_{i_1},x^{(2)}_{i_2},x^{(4)}_{i_4}) \\
    &= \frac {f^{(1)}_{i_1}f^{(2)}_{i_2}f^{(3)}_{c}f^{(4)}_{i_4}} {f^{(1)}_{i_1}f^{(2)}_{i_2}f^{(3)}_{c}f^{(4)}_{i_4}+(1-f^{(1)}_{i_1})(1-f^{(2)}_{i_2})(1-f^{(3)}_{c})(1-f^{(4)}_{i_4})} \\
    \end{split}
\end{equation}
\end{small}
\begin{small}
\begin{equation}\label{equ:f_joint_com3}
\begin{split}
    &\ Fjoint(x^{(1)}_{i_1},x^{(2)}_{i_2}) \\
    &= \frac {f^{(1)}_{i_1}f^{(2)}_{i_2}f^{(3)}_{c}f^{(4)}_{c}} {f^{(1)}_{i_1}f^{(2)}_{i_2}f^{(3)}_{c}f^{(4)}_{c}+(1-f^{(1)}_{i_1})(1-f^{(2)}_{i_2})(1-f^{(3)}_{c})(1-f^{(4)}_{c})} \\
    \end{split}
\end{equation}
\end{small}
where $f^{(3)}_{c} = \sqrt {Fcross^{(1)-(3)}_{i_1}Fcross^{(2)-(3)}_{i_2}}$ and $f^{(4)}_{c} = \sqrt {Fcross^{(1)-(4)}_{i_1}Fcross^{(2)-(4)}_{i_2}}$ are the completion views constructed from cross-view functions. The detailed view completion experiments are listed in Table \ref{tab:sim2}. Equipped with view completion, the clustering performance has been improved by about 0.6\%-0.8\%, proving the effectiveness of consistency learning and view completion.

\subsection{Ablation Studies And Parameter Analysis}
In this section, we conduct some studies on several datasets in the following. 

\begin{figure*}
    \centering
    \includegraphics[width=0.31\textwidth]{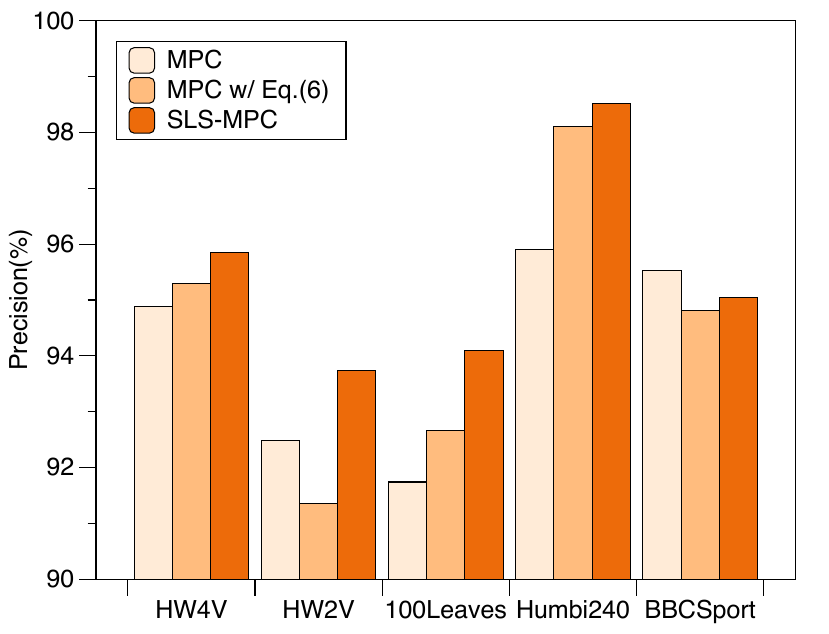}
    \includegraphics[width=0.31\textwidth]{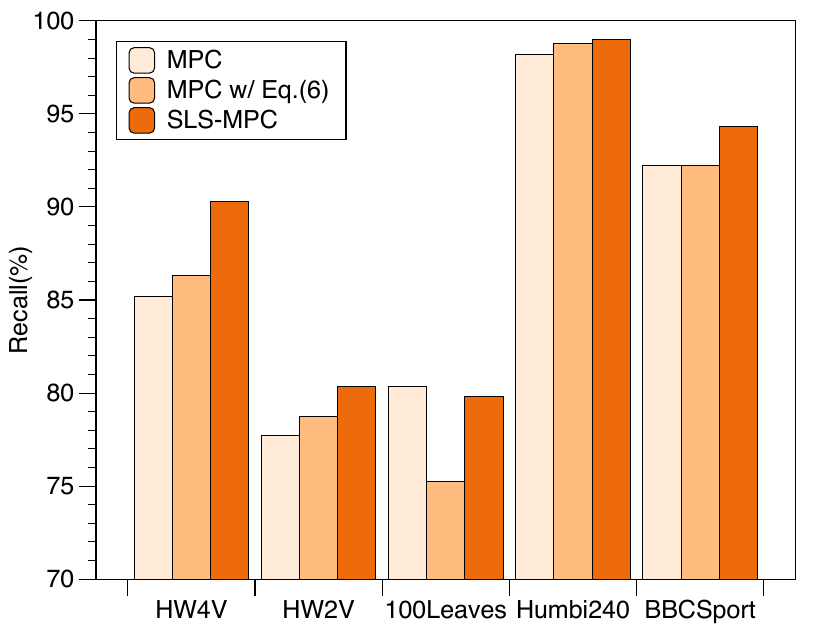}
    \includegraphics[width=0.31\textwidth]{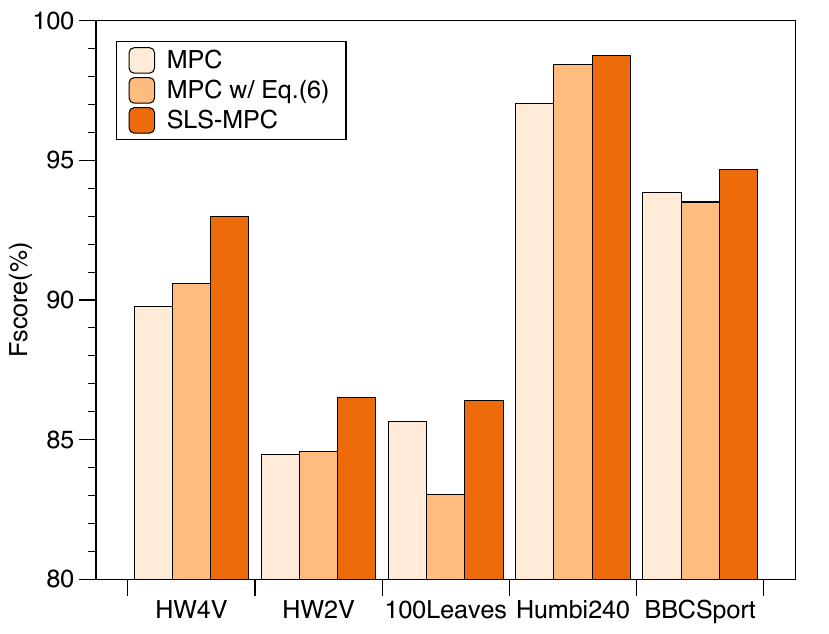}
    \caption{Ablation study of our method. Comparison on probability estimation between MPC, MPC w/ Eq. (\ref{equ:p_bayesian_new}) and SLS-MPC.}
    \label{fig:fg_estimation}
\end{figure*}

\noindent{\bf Ablation on Probability Estimation.}
In the probability estimation, we use Eq. (\ref{equ:p_bayesian_new}) to fuse the probability information of each view. In Table \ref{tab:ablation-formula}, we compare the formula with different aggregation functions on Handwritten with two views and four views. And the aggregation function is expressed as: $P(i,j)=Aggregation(P(e_{ij}=1|w^{(1)}),P(e_{ij}=1|w^{(2)}),...,P(e_{ij}=1|w^{(M)}))$, where aggregation functions include mean, max, min and multiply. The mean function treats multiple views as equally important and cannot generate good clustering result. Compared with the naive max function, SLS-MPC using the formula in Eq. (\ref{equ:p_bayesian_new}) can significantly boost the ARI from 78.25 to 92.56 on handwritten with four views. It further proves that Eq. (\ref{equ:p_bayesian_new}) can adaptively estimate the posterior matching probability from multiple views. From the perspective of multi-view probability estimation, we compare our method with MPC and MPC using Eq. (\ref{equ:p_bayesian_new}) in Fig. \ref{fig:fg_estimation}. The performance of MPC using Eq. (\ref{equ:p_bayesian_new}) is about 0.80\% higher than that of MPC on Handwritten with four views in terms of BCubed Fscore. And the performance of SLS-MPC is about 2.41\% higher than that of MPC using Eq. (\ref{equ:p_bayesian_new}) on Handwritten with four views in terms of BCubed Fscore. These experimental results prove that our formula proposed in Eq. (\ref{equ:p_bayesian_new}) can adaptively fuse multi-view probability information in an efficient way, which plays a major role in performance improvement.

\begin{table}[htbp]
  \centering
  \caption{Ablation study of our method. Comparison between the formula and the different aggregation functions on Humbi240 and Handwritten.}
    \begin{tabular}{|c|c|cccc|}
    \hline
    Datasets & Methods & FP    & FB    & NMI   & ARI \bigstrut\\
    \hline
    \multirow{5}[2]{*}{Humbi240} & max   & 87.67 & 89.67 & 96.39 & 87.62 \bigstrut[t]\\
          & mean  & 92.14 & 93.46 & 97.71 & 92.11 \\
          & min   & 97.2  & 98.04 & 99.37 & 97.19 \\
          & multiply & 97.26 & 98.03 & 99.35 & 97.25 \\
          & formula & \textbf{98.12} & \textbf{98.77} & \textbf{99.62} & \textbf{98.11} \bigstrut[b]\\
    \hline
    \multirow{5}[2]{*}{\tabincell{c}{Handwritten \\ view 1-2}} & max   & 73.86 & 74.33 & 80.18 & 71.45 \bigstrut[t]\\
          & mean  & 80.6  & 80.35 & 83.52 & 78.79 \\
          & min   & 83.83 & 82.99 & 84.42 & 82.24 \\
          & multiply & 84.16 & 83.45 & 85.01 & 82.61 \\
          & formula & \textbf{87.03} & \textbf{86.51} & \textbf{87.62} & \textbf{85.73} \bigstrut[b]\\
    \hline
    \multirow{5}[2]{*}{\tabincell{c}{Handwritten \\ view 1-4}} & max   & 80.15 & 79.76 & 83.56 & 78.25 \bigstrut[t]\\
          & mean  & 88.96 & 88.53 & 88.46 & 87.83 \\
          & min   & 87.21 & 86.64 & 86.80 & 85.90 \\
          & multiply & 90.75 & 90.23 & 89.69 & 89.78 \\
          & formula & \textbf{93.28} & \textbf{92.99} & \textbf{92.13} & \textbf{92.56} \bigstrut[b]\\
    \hline
    \end{tabular}%
  \label{tab:ablation-formula}%
\end{table}%

\begin{table}[htbp]
  \centering
  \caption{Ablation study of our method. Comparison between different similarity measures in the complete cases.}
    \begin{tabular}{|c|c|cccc|}
    \hline
    Datasets & Methods & FP    & FB    & NMI   & ARI \bigstrut\\
    \hline
    \multirow{4}[2]{*}{100Leaves} & $L_1$    & \textbf{89.65 } & \textbf{90.60 } & \textbf{96.49 } & \textbf{89.56 } \bigstrut[t]\\
          & $L_2$    & 84.35  & 85.70  & 94.83  & 84.22  \\
          & $L_3$    & 79.35  & 80.64  & 92.96  & 79.18  \\
          & Cosine & 85.46  & 86.39  & 95.03  & 85.34  \bigstrut[b]\\
    \hline
    \multirow{4}[2]{*}{Humbi240} & $L_1$     & 96.94  & 98.09  & 99.44  & 96.93  \bigstrut[t]\\
          & $L_2$    & 97.75  & 98.60  & 99.59  & 97.74  \\
          & $L_3$   & 97.79  & 98.63  & 99.60  & 97.78  \\
          & Cosine & \textbf{98.12 } & \textbf{98.77 } & \textbf{99.62 } & \textbf{98.11 } \bigstrut[b]\\
    \hline
    \multirow{4}[2]{*}{\tabincell{c}{Handwritten \\ view 1-2}} & $L_1$    & 86.08  & 85.55  & 87.04  & 84.69  \bigstrut[t]\\
          & $L_2$    & 86.91  & 86.44  & 87.47  & 85.59  \\
          & $L_3$    & \textbf{87.07 } & \textbf{86.69 } & \textbf{87.66 } & \textbf{85.76 } \\
          & Cosine & 87.03  & 86.51  & 87.62  & 85.73  \bigstrut[b]\\
    \hline
    \multirow{4}[2]{*}{\tabincell{c}{Handwritten \\ view 1-4}} & $L_1$    & \textbf{93.32 } & \textbf{93.03 } & \textbf{92.15 } & \textbf{92.60 } \bigstrut[t]\\
          & $L_2$    & 92.32  & 92.06  & 91.35  & 91.51  \\
          & $L_3$    & 91.76  & 91.49  & 91.18  & 90.89  \\
          & Cosine & 93.28  & 92.99  & 92.13  & 92.56  \bigstrut[b]\\
    \hline
    \end{tabular}%
  \label{tab:sim}%
\end{table}%

\begin{table}[htbp]
  \centering
  \caption{Ablation study of our method. Comparison between different similarity measures in the incomplete cases. VP indicates view completion proposed in Eq. (\ref{equ:f_joint_com1}), Eq. (\ref{equ:f_joint_com2}) and Eq. (\ref{equ:f_joint_com3}).}
    \begin{tabular}{|c|c|cccc|}
    \hline
    Datasets & Methods & FP    & FB    & NMI   & ARI \bigstrut\\
    \hline
    \multirow{4}[2]{*}{\tabincell{c}{Handwritten \\ view 1-4}} & $L_1$      & 89.06          & 88.97          & 89.06          & 87.94  \bigstrut[t]\\
          & $L_2$      & 88.17          & 88.18          & 88.36          & 86.97  \\
          & $L_3$      & 89.47          & 89.15          & 89.04          & 88.39 \\
          & Cosine  & \textbf{91.73} & \textbf{91.42} & \textbf{90.92} & \textbf{90.86}  \bigstrut[b]\\
    \hline
    \multirow{4}[2]{*}{\tabincell{c}{Handwritten \\ view 1-4 VP}} & $L_1$    & 90.04          & 89.95          & 89.83          & 89.01 \bigstrut[t]\\
          & $L_2$      & 89.51          & 89.45          & 89.29          & 88.44  \\
          & $L_3$      & 90.32          & 90.04          & 89.81          & 89.33  \\
          & Cosine  & \textbf{92.48} & \textbf{92.18} & \textbf{91.56} & \textbf{91.69}  \bigstrut[b]\\
    \hline
    \end{tabular}%
  \label{tab:sim2}%
\end{table}%

\noindent{\bf{Ablation on Similarity Measures.}} To keep consistent with previous works MPC\cite{MPC}, PIC\cite{PIC} and UEAF\cite{UEAF}, we use cosine metric to estimate the similarity matrix. As listed in Table \ref{tab:sim} and Table \ref{tab:sim2}, we report the clustering performance in complete cases and incomplete cases obtained using similarity metric $L_p$, where $L_p(x_i, x_j) = (\sum_{l=1}^{n} {|x_i^{(l)}-x_j^{(l)}|^p})^{\frac {1} {p}},x_i=(x_i^{(1)},...,x_i^{(n)})$. Overall, SLS-MPC is robust to the choice of metric and the performance using cosine metric is more stable than that of $L_p$.

\begin{table}[htbp]
  \centering
  \caption{Ablation study of our method. Comparison on Consistency Loss.}
    \begin{tabular}{|c|c|cccc|}
    \hline
    Dataset & Consistency Loss & FP    & FB    & NMI   & ARI \bigstrut\\
    \hline
    \multirow{2}[2]{*}{\tabincell{c}{Handwritten \\ view 1-2}} & w/ Eq. (\ref{equ:loss_fz}) & 77.82  & 77.95  & 82.64  & 75.66  \bigstrut[t]\\
          & w/ Eq. (\ref{equ:loss_1})  & \textbf{87.03} & \textbf{86.51} & \textbf{87.62} & \textbf{85.73}  \bigstrut[b]\\
    \hline
    \multirow{2}[2]{*}{\tabincell{c}{Handwritten \\ view 1-4}} & w/ Eq. (\ref{equ:loss_fz}) & 80.43  & 80.34  & 82.41  & 78.71  \bigstrut[t]\\
          & w/ Eq. (\ref{equ:loss_1})  & \textbf{93.28} & \textbf{92.99} & \textbf{92.13} & \textbf{92.56}  \bigstrut[b]\\
    \hline
    \end{tabular}%
  \label{tab:loss2}%
\end{table}%

\begin{table}[htbp]
  \centering
  \caption{Ablation study of our method. Comparison on loss component.}
    \begin{tabular}{|c|c|cccc|}
    \hline
    Dataset & Component & FP    & FB    & NMI   & ARI \bigstrut\\
    \hline
    \multirow{4}[2]{*}{\tabincell{c}{Handwritten \\ view 1-2}} & w/o $L_{consistency1}$ & 83.63  & 83.00  & 84.60  & 82.03  \bigstrut[t]\\
          & w/o $L_{consistency2}$ & 86.15  & 85.95  & 87.39  & 84.77  \\
          & w/o $L_{constraint}$ & 80.81  & 80.41  & 83.97  & 79.06  \\
          & SLS-MPC  & \textbf{87.03} & \textbf{86.51} & \textbf{87.62} & \textbf{85.73}  \bigstrut[b]\\
    \hline
    \multirow{4}[2]{*}{\tabincell{c}{Handwritten \\ view 1-4}} & w/o $L_{consistency1}$ & 90.13  & 89.74  & 89.63  & 89.12  \bigstrut[t]\\
          & w/o $L_{consistency2}$ & 85.61  & 85.44  & 86.17  & 84.25  \\
          & w/o $L_{constraint}$ & 83.50  & 83.44  & 84.75  & 81.98  \\
          & SLS-MPC  & \textbf{93.28} & \textbf{92.99} & \textbf{92.13} & \textbf{92.56}  \bigstrut[b]\\
    \hline
    \end{tabular}%
  \label{tab:loss}%
\end{table}%

\begin{figure}
    \centering
    \includegraphics[width=0.24\textwidth]{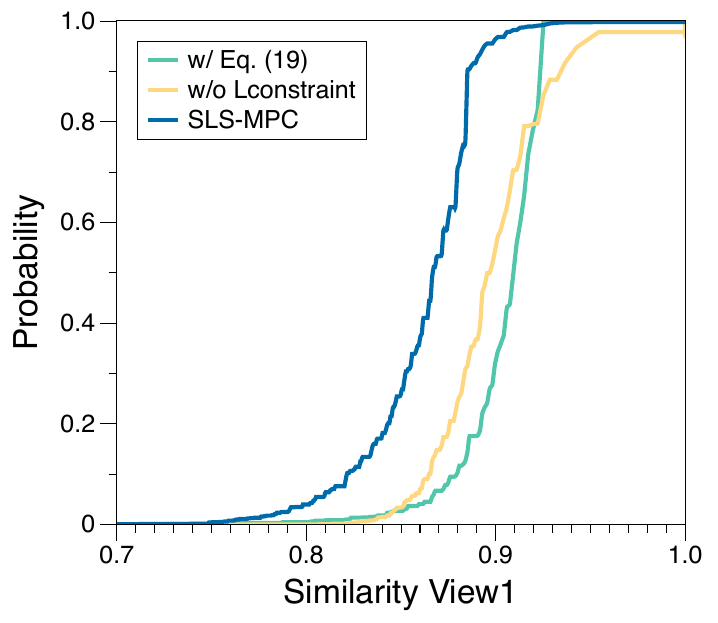}
    \includegraphics[width=0.24\textwidth]{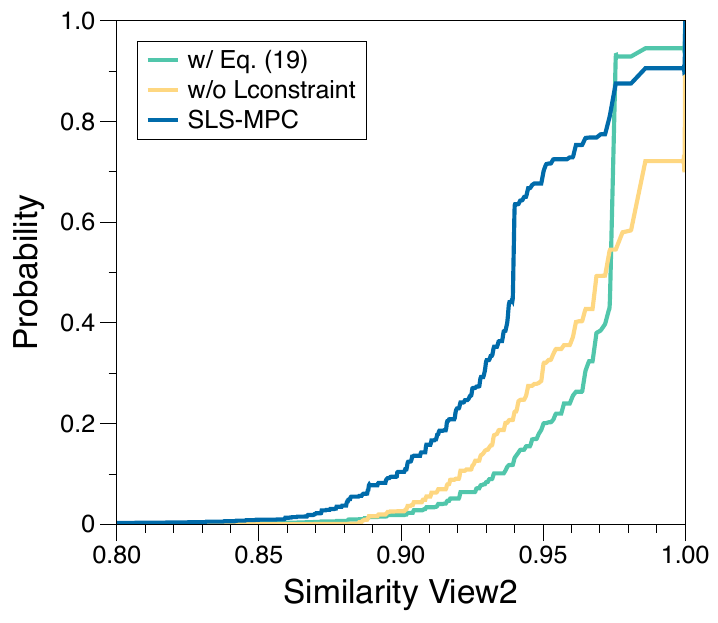}
    \includegraphics[width=0.24\textwidth]{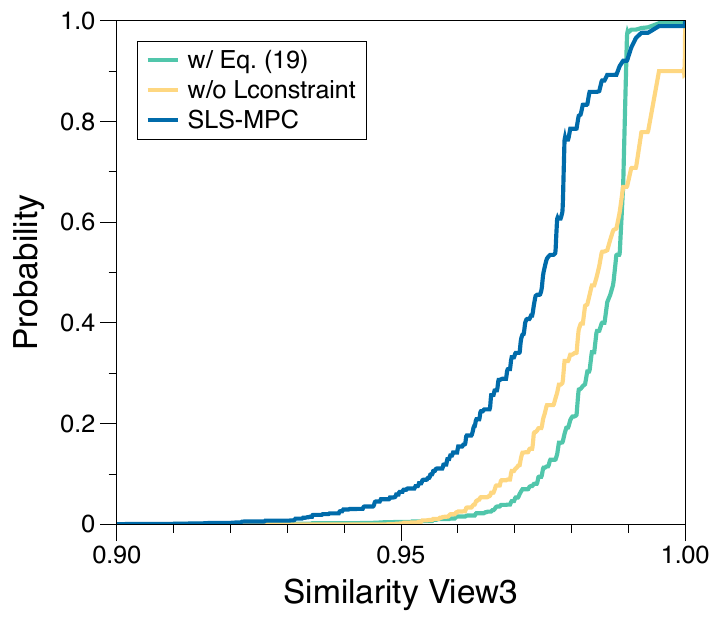}
    \includegraphics[width=0.24\textwidth]{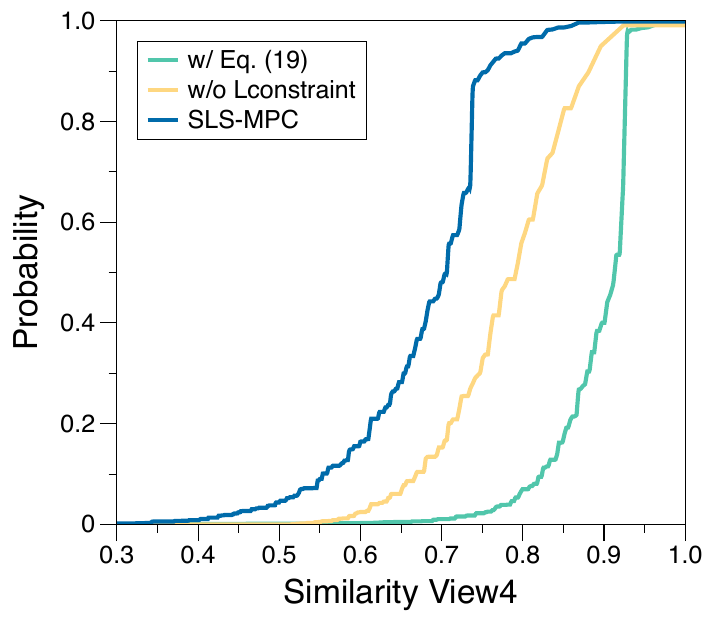}
    \caption{The visualization of self-learning probability function in Handwritten with four views.}
    \label{fig:fg_plt}
\end{figure}

\noindent{\bf{Ablation on Consistency Loss.}} As described in Section Self-Learning Probability Function, consistency loss Eq. (\ref{equ:loss_fz}) and Eq. (\ref{equ:loss_1}) are introduced in self-learning to learn probability function. As shown in Table \ref{tab:loss2}, using Eq. (\ref{equ:loss_fz}) results in poor clustering performance, which demonstrates that $Fsingle$ is confused by $Fmulti$ and $Fcross$ in the consistency learning process and the successful introduction of Eq. (\ref{equ:loss_1}) enables the learning of a better probability function. Moreover, as shown in Fig. \ref{fig:fg_plt}, using Eq. (\ref{equ:loss_fz}) causes the probability function to shift to the right. The probability function is relatively steep and the value of the probability function is low and inaccurate. Specifically, in the fourth view, the value of the probability function reaches 1.0 only when the similarity arrives at about 0.92. And, the value of the probability function varies greatly when the similarity fluctuates around 0.9.

\noindent{\bf{Ablation on Loss Component.}} As described in Eq. (\ref{equ:loss_all}), consistency loss and constraint loss are introduced in self-learning to learn probability function. As shown in Table \ref{tab:loss}, all loss terms play indispensable roles in SLS-MPC. Moreover, as shown in Fig. \ref{fig:fg_plt}, optimizing without $L_{constraint}$ makes the range of the probability function unconstrained. The maximum value of the probability function is about 0.7 and 0.9 in the second view and the third view respectively. It should be pointed out that optimizing without $L_{constraint}$ results in poor clustering performance, which demonstrates the importance of range constraint.

\begin{figure}
    \centering
    \includegraphics[width=0.45\textwidth]{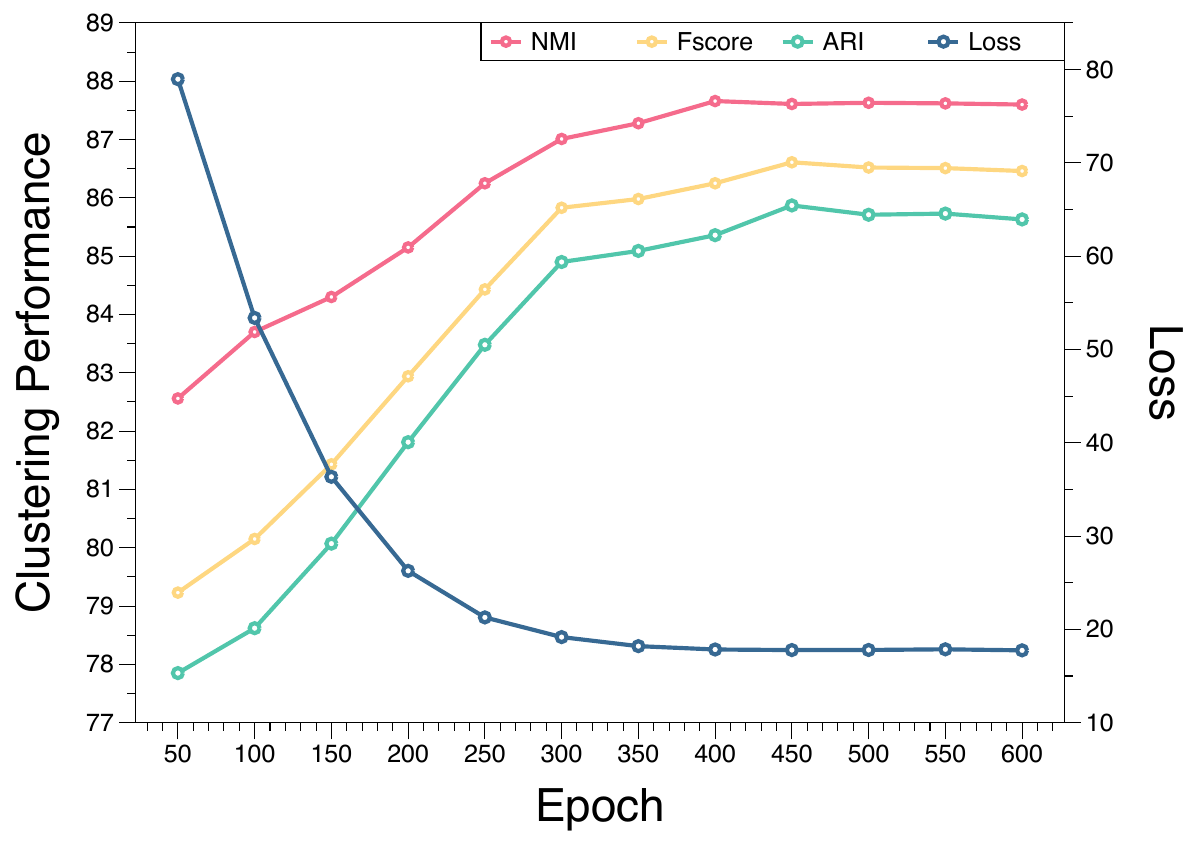}
    \caption{The clustering performance of SLS-MPC with increasing epoch on Handwritten. The x-axis denotes the epoch in iteration, the left and right y-axis denote the clustering performance and corresponding loss value, respectively.}
    \label{fig:epoch}
\end{figure}

\begin{figure}
    \centering
    \includegraphics[width=0.4\textwidth]{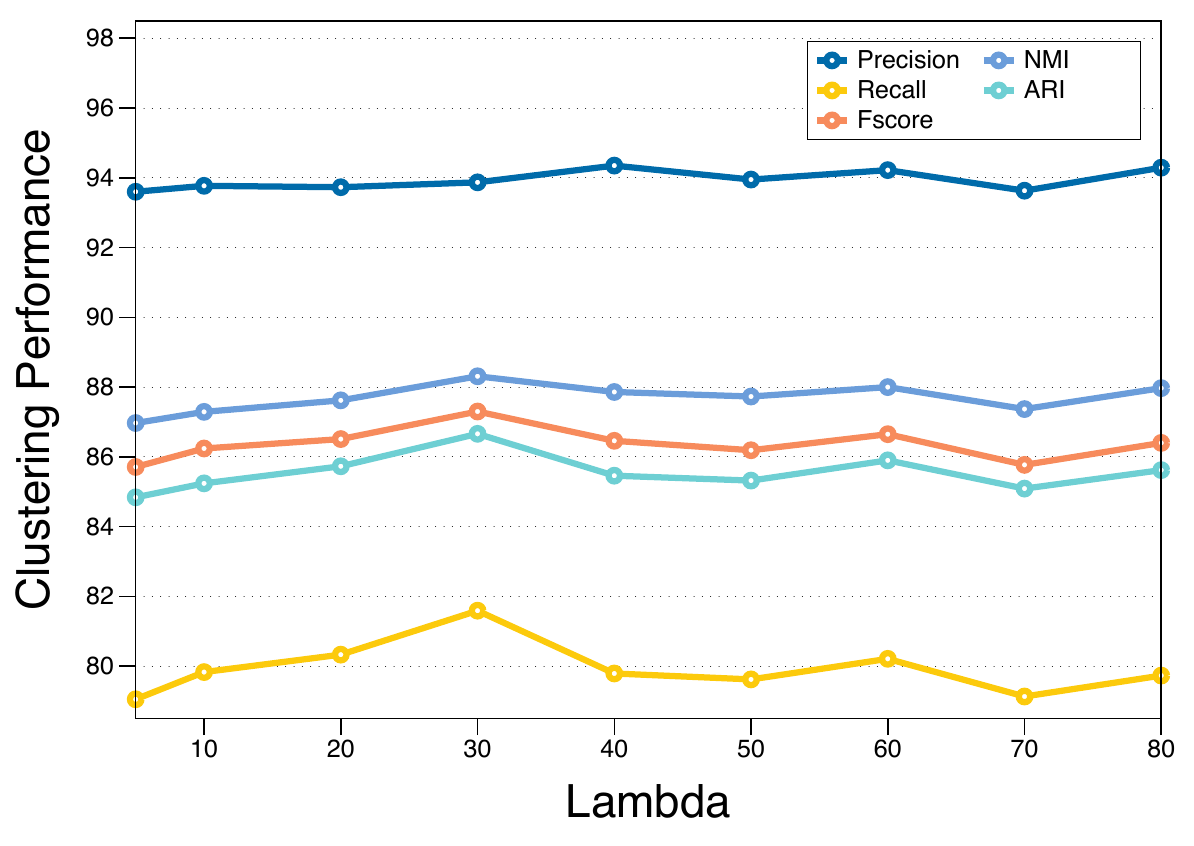}
    \caption{The analysis of parameter $\lambda$ on Handwritten.}
    \label{fig:lam}
\end{figure}

\noindent{\bf{Analysis of Convergence.}}
In this sub-section, we analyze the convergence of SLS-MPC by reporting the loss value and the corresponding clustering performance with increasing epochs. As shown in Fig. \ref{fig:epoch}, the loss value remarkably decreases in the first 300 epochs, and meanwhile NMI, Fscore, and ARI continuously increase. And then the clustering performance keeps stable in the last moving epochs.

\noindent{\bf{Analysis of Parameter $\lambda$.}}
According to Eq. (\ref{equ:loss_all}), objective function contains a balanced factor $\lambda$ on $L_{consistency}$ and $L_{constraint}$. We choose nine values from 5 to 80 to study how it affects the clustering performance on Handwritten with two views. As shown in Fig. \ref{fig:lam}, the clustering performance is robust when the factor $\lambda$ changes and precision is stable when the factor $\lambda$ is around 20, which is the value we used for reporting performance in the above results on Handwritten with two views. The detailed factor $\lambda$ used in our experiments is listed in Table \ref {tab:setting}.

\begin{table}[htbp]
  \centering
  \caption{The clustering performance of EEIMC and PIC with MPC and SLS-MPC.}
    \begin{tabular}{|c|ccc|}
    \hline
    Methods & Handwritten & 100Leaves & Humbi240 \bigstrut\\
    \hline
    EEIMC\cite{EEIMC} & 76.51  & 73.84  & 91.41  \bigstrut[t]\\
    EEIMC w/ MPC & +9.69  & +11.63  & -0.70  \\
    EEIMC w/ SLS-MPC & \textbf{+10.81 } & \textbf{+14.73 } & \textbf{+0.48 } \bigstrut[b]\\
    \hline
    PIC\cite{PIC}   & 73.94  & 77.82  & 94.32  \bigstrut[t]\\
    PIC w/ MPC & +14.87  & +8.78  & +1.62  \\
    PIC w/ SLS-MPC & \textbf{+17.24 } & \textbf{+12.66 } & \textbf{+2.62 } \bigstrut[b]\\
    \hline
    \end{tabular}%
  \label{tab:mp}%
\end{table}%

\noindent{\bf{Analysis of Multi-view Probability.}}
We use multi-view probability generated from MPC and our proposed SLS-MPC to replace the kernel matrix in EEIMC\cite{EEIMC} and the similarity matrix in PIC\cite{PIC}. The clustering results are listed in Table \ref{tab:mp}. Compared with origin kernel matrix and similarity matrix, the performance of EEIMC and PIC using multi-view probability are improved which further demonstrates that the accuracy of multi-view probability is better than that of origin similarity and using SLS-MPC works better which demonstrates the effectiveness of symmetry and self-learning in SLS-MPC.

\section{Conclusion}
\label{sec:conclusion}
In this paper, we propose self-learning symmetric multi-view probabilistic clustering (SLS-MPC) to tackle the challenges: i) lack of unified framework for incomplete and complete MVC, ii) lack of emphasis on noise and outliers and iii) dependence on category information and complex hyper-parameters. SLS-MPC proposes a novel self-learning probability function to effectively learn each view's individual distribution without any prior knowledge and hyper-parameters from the aspect of consistency in single-view, cross-view and multi-view and a novel method to adaptively estimate the posterior matching probability from multiple views without complicated hyper-parameters fine-tuning, which tolerates incomplete views. Besides, equipped with graph-context-aware probability refinement, SLS-MPC takes noise and outliers into consideration. Moreover, SLS-MPC proposes a novel probabilistic clustering algorithm, which has no optimization parameters and generates clustering results in an unsupervised manner and an efficient way without category information. Extensive experiments on multiple benchmarks for incomplete and complete MVC show that our proposed SLS-MPC performs markedly better than SOTA methods.

\section*{Acknowledgments}
\label{sec:ack}
This work was supported in part by Zhejiang Provincial Natural Science Foundation of China under Grant No. LDT23F01013F01; in part by the Fundamental Research Funds for the Central Universities; in part by Alibaba Group through Alibaba Research Intern Program.


\bibliographystyle{IEEEtran}
\bibliography{IEEEabrv,Sample}

\vspace{-33pt}

\begin{IEEEbiography}
[{\includegraphics[width=1in,height=1.25in,clip,keepaspectratio]{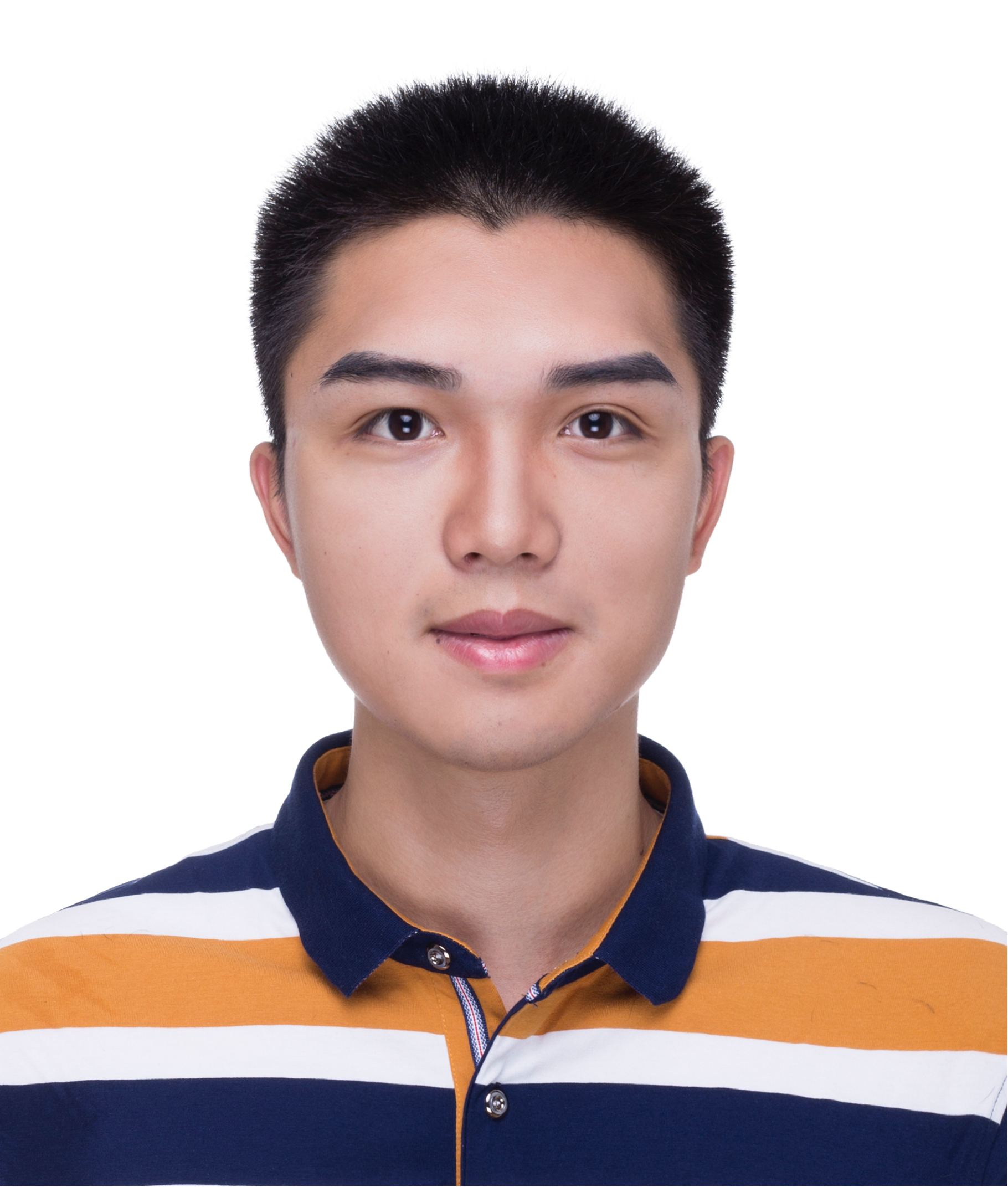}}]{Junjie Liu}
was born in Jiangsu Province, China, in 1995. He received the B.Sc. and Ph.D. degrees from Zhejiang University, Hangzhou, China, in 2017 and 2024. He is currently an algorithm engineer at Alibaba Cloud, Hangzhou, China. His research interests include image processing, computer vision and machine learning.
\end{IEEEbiography}
\vspace{-33pt}

\begin{IEEEbiography}[{\includegraphics[width=1in,height=1.25in,clip,keepaspectratio]{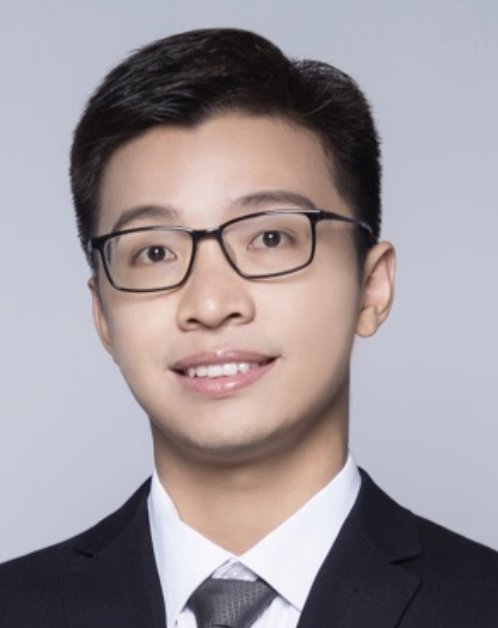}}]{Junlong Liu}
received the B.S. degree in computer science from Beihang University and the M.S. degree in machine learning from University of Science and Technology of China, China, in 2014 and 2017. He is currently an algorithm engineer at Alibaba Cloud, Hangzhou, China. His research interests include image processing, computer vision and machine learning.
\end{IEEEbiography}
\vspace{-33pt}

\begin{IEEEbiography}[{\includegraphics[width=1in,height=1.25in,clip,keepaspectratio]{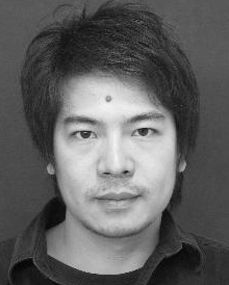}}]{Rongxin Jiang}
was born in Hunan Province, China, in 1982. He received the B.Sc. and Ph.D. degrees in computer vision from Zhejiang University, Hangzhou, China, in 2002 and 2008, respectively. He is currently an Associate Professor of Zhejiang University. His major research fields are computer vision and networking.
\end{IEEEbiography}
\vspace{-33pt}

\begin{IEEEbiography}[{\includegraphics[width=1in,height=1.25in,clip,keepaspectratio]{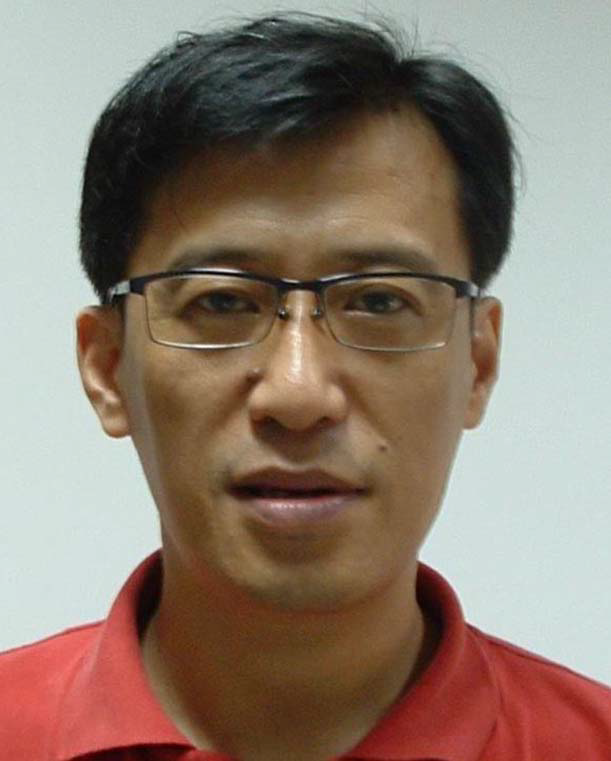}}]{Yaowu Chen}
was born in Heilongjiang Province, China, in 1963. He received the Ph.D. degree in embedded system from Zhejiang University, Hangzhou, China, in 1998. He is currently a Professor and the Director of the Institute of Advanced Digital Technologies and Instrumentation, Zhejiang University. His major research fields are embedded system, multimedia system, and networking.
\end{IEEEbiography}
\vspace{-33pt}

\begin{IEEEbiography}[{\includegraphics[width=1in,height=1.25in,clip,keepaspectratio]{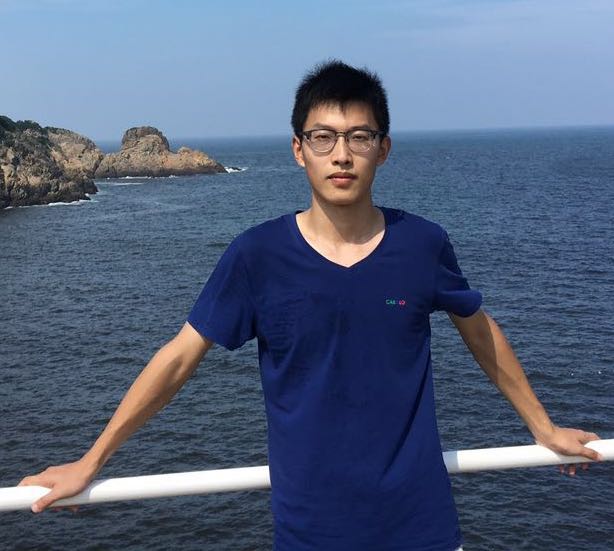}}]{Chen Shen}
received his B.S. degree and Ph.D. in Electrical Engineering at Zhejiang University, China, in 2012 and 2018. Now he is a Senior algorithm Engineer at Alibaba Cloud, Hangzhou, China. His research interests include deep learning, data mining and large language models.
\end{IEEEbiography}
\vspace{-33pt}

\begin{IEEEbiography}[{\includegraphics[width=1in,height=1.25in,clip,keepaspectratio]{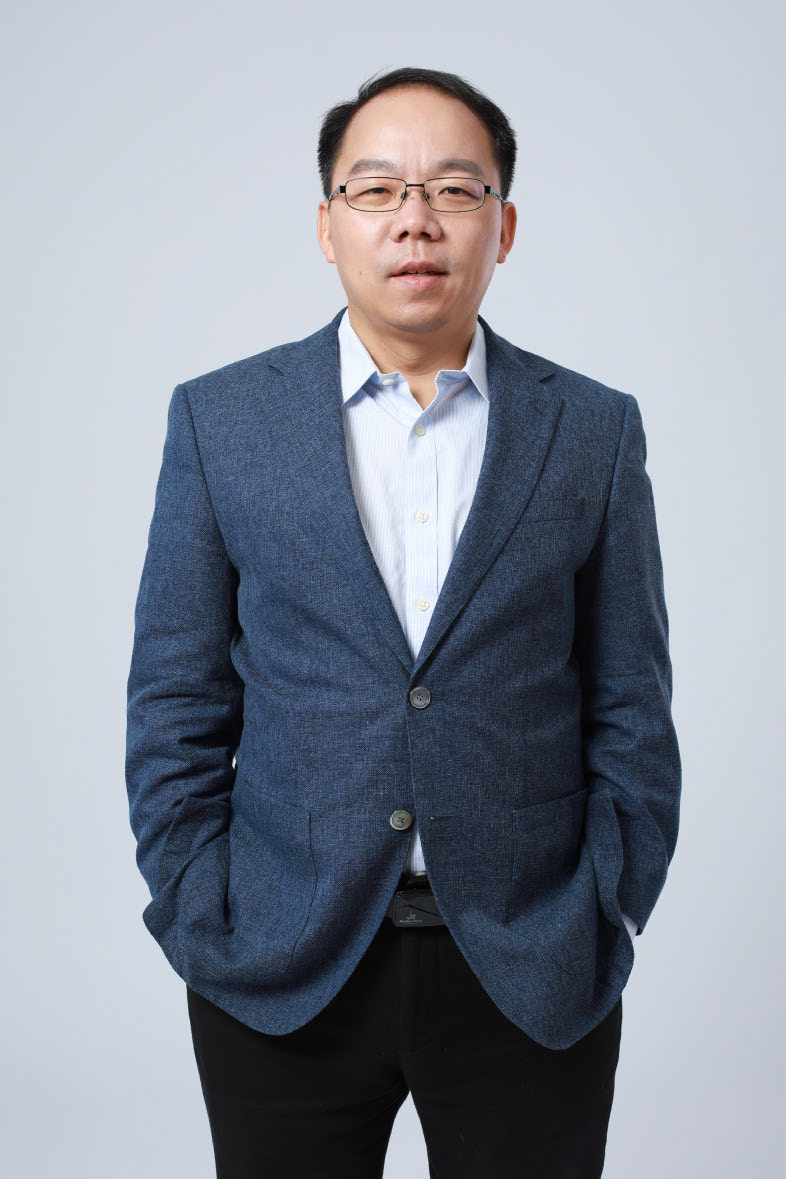}}]{Jieping Ye}
is a VP of Alibaba Cloud. His research interests include big data, machine learning, and artificial intelligence with applications in transportation, smart city, and biomedicine. He has served as a Senior Program Committee/Area Chair/Program Committee Vice Chair of many conferences including NeurlPS, ICML, KDD, IJCAI, ICDM, and SDM. He has served as an Associate Editor of Data Mining and Knowledge Discovery, IEEE Transactions on Knowledge and Data Engineering, and IEEE Transactions on Pattern Analysis and Machine Intelligence. He won the NSF CAREER Award in 2010. His papers have been selected for the outstanding student paper at ICML in 2004, the KDD best research paper runner up in 2013, and the KDD best student paper award in 2014. He has also won the first place in 2019 INFORMS Daniel H. Wagner Prize, one of the top awards in operation research practice. Dr. Ye was elevated to an IEEE Fellow in 2019 and named an ACM Distinguished Scientist in 2020 for his contributions to the methodolog
\end{IEEEbiography}
\vspace{-33pt}


\vfill

\end{document}